\documentclass[10pt,twocolumn,letterpaper]{article}

\usepackage[iccvfinal]{iccv}      

\usepackage{graphicx}	
\usepackage{cuted}
\usepackage{makecell}	
\usepackage{amssymb}	
\usepackage{booktabs}
\usepackage{times}
\usepackage{booktabs}
\usepackage{xcolor}
\usepackage{wrapfig}
\usepackage{colortbl}
\usepackage{makecell}
\usepackage{multirow}
\usepackage{amssymb}
\definecolor{lightblue}{RGB}{180, 210, 255}
\usepackage{bbding}
\usepackage{pifont}

\usepackage{amsmath}
\usepackage{microtype}
\usepackage{epsfig}
\usepackage{caption}
\usepackage{float}
\usepackage{placeins}
\usepackage{stfloats}
\usepackage{enumitem}
\usepackage{tabularx}
\usepackage{xstring}
\usepackage{xspace}
\usepackage{subcaption}
\usepackage{xcolor}

\definecolor{iccvblue}{rgb}{0.21,0.49,0.74}
\usepackage[pagebackref,breaklinks,colorlinks,allcolors=iccvblue]{hyperref}

\def\paperID{*****} 
\def\confName{ICCV}
\def\confYear{2025}

\title{FICGen: Frequency-Inspired Contextual Disentanglement for \\Layout-driven Degraded Image Generation}

\author{Wenzhuang Wang\textsuperscript{\rm 1,\rm 2}, Yifan Zhao\textsuperscript{\rm 1}\thanks{Correspondence should be addressed to Yifan Zhao and Jia Li.}, Mingcan Ma\textsuperscript{\rm 2}, Ming Liu\textsuperscript{\rm 2}, Zhonglin Jiang\textsuperscript{\rm 2}, Yong Chen\textsuperscript{\rm 2}, Jia Li\textsuperscript{\rm 1}\footnotemark[1] \\
\textsuperscript{\rm 1}State Key Laboratory of Virtual Reality Technology and Systems, SCSE\&QRI, Beihang University\\
\textsuperscript{\rm 2}Geely Automobile Research Institute (Ningbo) Co., Ltd\\
{\tt\small \{wz\_wang,zhaoyf,jiali\}@buaa.edu.cn}
}

\begin{document}
\maketitle
\begin{strip}
\begin{minipage}{\textwidth}\centering
\vspace{-9mm}
\includegraphics[width=0.96\textwidth]{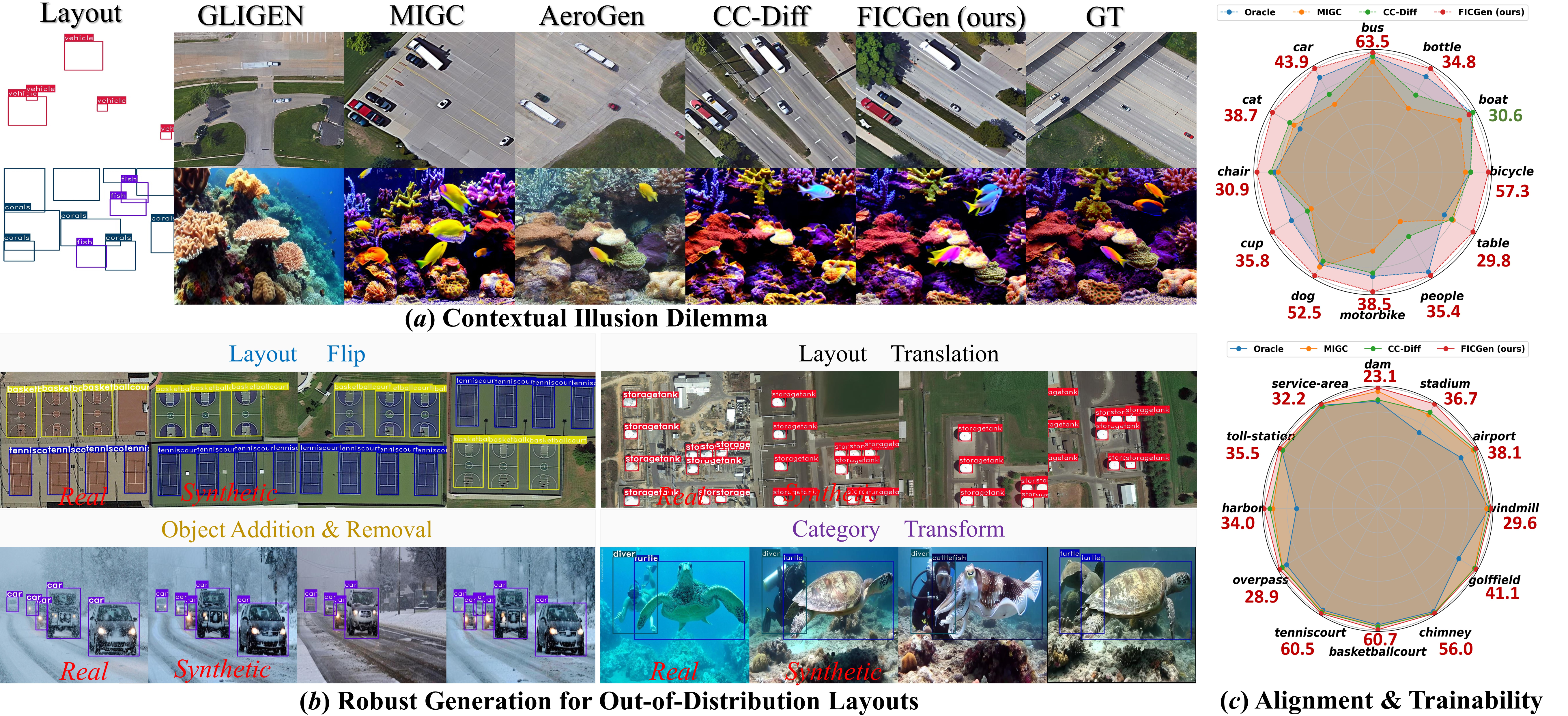}
\captionof{figure}{\textbf{Visualizations of degraded image generation by our FICGen.} FICGen achieves superior quality compared to existing L2I methods and demonstrates strong robustness to OOD layouts. Moreover, it shows excellent trainability for boosting object detectors, \textit{i.e.,} Faster R-CNN (R50) \cite{ren2015faster}. The alignment and trainability are evaluated on the ExDARK \cite{loh2019getting} and DIOR-H~\cite{li2020object}, respectively.}
\label{fig1}
\end{minipage}
\end{strip}
\begin{abstract}
Layout-to-image (L2I) generation has exhibited promising results in natural domains, but suffers from limited generative fidelity and weak alignment with user-provided layouts when applied to degraded scenes (\textit{i.e.}, low-light, underwater). We primarily attribute these limitations to the ``\textbf{contextual illusion dilemma}'' in degraded conditions, where foreground instances are overwhelmed by context-dominant frequency distributions. Motivated by this, our paper proposes a new \textbf{F}requency-\textbf{I}nspired \textbf{C}ontextual Disentanglement \textbf{Gen}erative (FICGen) paradigm, which seeks to transfer frequency knowledge of degraded images into the latent diffusion space, thereby facilitating the rendering of degraded instances and their surroundings via contextual frequency-aware guidance. To be specific, FICGen consists of two major steps. Firstly, we introduce a learnable dual-query mechanism, each paired with a dedicated frequency resampler, to extract contextual frequency prototypes from pre-collected degraded exemplars in the training set. Secondly, a visual-frequency enhanced attention is employed to inject frequency prototypes into the degraded generation process. To alleviate the contextual illusion and attribute leakage, an instance coherence map is developed to regulate latent-space disentanglement between individual instances and their surroundings, coupled with an adaptive spatial-frequency aggregation module to reconstruct spatial-frequency mixed degraded representations. Extensive experiments on \textbf{5} benchmarks involving a variety of degraded scenarios—\textbf{from severe low-light to mild blur}—demonstrate that FICGen consistently surpasses existing L2I methods in terms of generative fidelity, alignment and downstream auxiliary trainability. The code is available at \href{https://github.com/fayewong666999/FICGen}{URL}.
\end{abstract}    
\section{Introduction}
\label{sec:intro}
Visual perception~\cite{ma2023boosting, shan2025ros} depends on a large amount of annotated datasets, particularly in degraded conditions such as low-light~\cite{loh2019getting}, and adverse weather~\cite{kenk2020dawn}.  Unfortunately, acquiring adequate datasets for these extreme conditions remains a major obstacle due to the high cost of manual collection and annotation. For instance, the number of images in ExDARK (7,363)~\cite{loh2019getting} is only about 1/20th that of COCO 2017 (118,287) ~\cite{lin2014microsoft}, highlighting the severe data scarcity. In light of this, alternative strategies are emerging that shift the focus from a ``model-centric'' to a ``data-centric'' perspective, where advanced generative models are used to synthesize realistic images for bolstering downstream tasks~\cite{wang2025freegen}.

Witnessing the phenomenal capabilities of text-to-image (T2I) diffusion models~\cite{ramesh2022hierarchical,saharia2022photorealistic}, numerous advances have been explored in conditioning image generation on visual labels, thereby enabling fine-grained controllability beyond what textual prompts alone can achieve. Such labels are often represented by masks~\cite{xue2023freestyle, lv2024place}, depths~\cite{zhang2023jointnet}, and bounding boxes (bboxes)~\cite{li2023gligen, yang2023reco, zheng2023layoutdiffusion, zhou2024migc, gu2024roictrl}, thereby facilitating visual generation via layout-to-image (L2I) synthesis. Among them, bboxes are the most intuitive for guiding the location of visual entities during L2I generation. 

Despite the success of L2I methods~\cite{li2023gligen, zhou2024migc, zheng2023layoutdiffusion} in natural scenarios, they face challenges when adapting to practical degraded conditions. In Fig.~\ref{fig1} (a), aerial objects (\ie, vehicles) are often small in size and visually resemble surrounding structures (\ie, bridges), while underwater species (\ie, fish) frequently blend with other nearby creatures (\ie, corals), resulting in contextual entanglement. Such cases, difficult even for the human eye to distinguish and predominantly occurring in degraded scenarios, lead to generative illusions regarding object quantity, location, and interaction—termed the \textbf{``contextual illusion dilemma''} in our work. 

\begin{figure}
\centering
\includegraphics[width=0.48\textwidth]{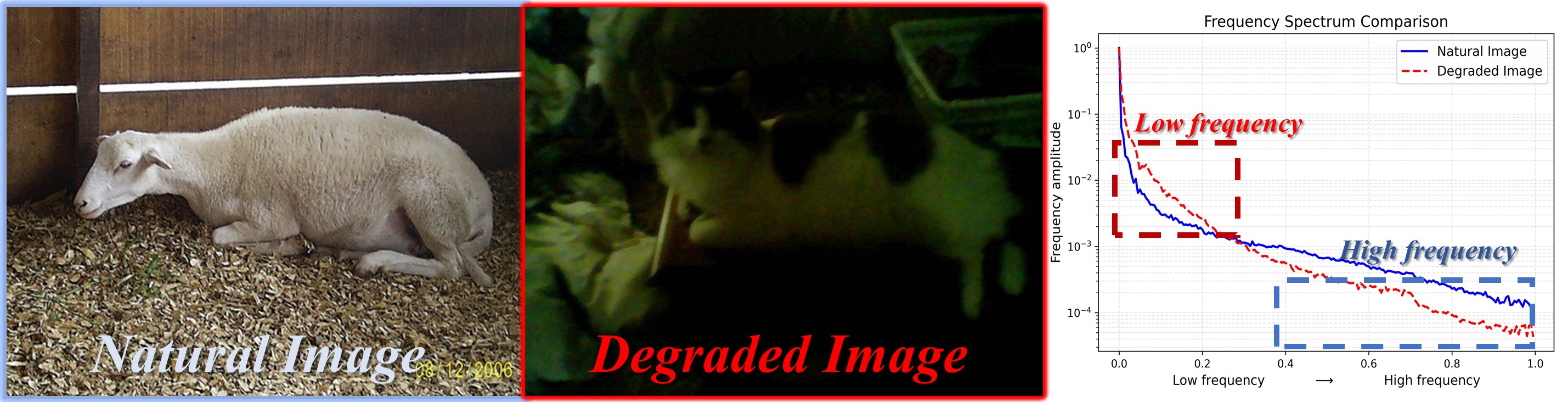}
\caption{Frequency spectrum comparison between natural and degraded images.}
\label{figattributes}
\end{figure}

Generally, as shown in Fig.~\ref{figattributes}, natural image exhibits a relatively balanced distribution of high-frequency (HF) and low-frequency (LF) components, making the distinction between instances and contexts pronounced. In contrast, HF details of instances in degraded contexts are attenuated, while LF contexts dominate the overall frequency distribution. Motivated by the frequency properties, we propose a new frequency-inspired contextual disentanglement paradigm to tackle the dilemma. Specifically, we first extract HF and LF prototypes from degraded contexts and associate them with each object and its surroundings. Second, a learnable dual-query mechanism equipped with frequency perceiver resamplers is introduced to simultaneously capture contextual frequency knowledge (\ie, boundaries and colours). Third, we inject the knowledge into the latent diffusion space via a visual-frequency enhanced attention, enabling better perception of frequency attributes during degraded generation. Meanwhile, an instance coherence map is designed to disentangle degraded instances from their surroundings, mitigating contextual entanglement and attribute leakage. Finally, a spatial-frequency aggregation module is developed to adaptively recombine them to obtain refined degraded representations. With the above efforts, FICGen offers a unified approach for degraded image synthesis from a frequency-inspired perspective. 
We evaluate FICGen on \textbf{5} degraded benchmarks, covering low-light (ExDARK~\cite{loh2019getting}), underwater (RUOD~\cite{fu2023rethinking}), remote sensing (DIOR~\cite{li2020object}), adverse weather (DAWN~\cite{kenk2020dawn}), and blur (VOC 2012~\cite{everingham2010pascal}) conditions. Extensive experiments demonstrate that FICGen achieves superior image quality while maintaining strong alignment with geometric layouts. In Fig.\ref{fig1}(b), FICGen can perform robust OOD generation on unseen layouts. Fig.\ref{fig1}(c) shows that FICGen achieves better alignment than the real test set across all classes except ``\textit{boat}'', effectively narrowing the real-synthetic domain gap. Furthermore, FICGen consistently improves downstream tasks by $\sim$\textbf{2.0} mAP and substantially boosts detection accuracy for certain classes, \ie, $\sim$\textbf{6.0 (32.2 \textit{vs.} 38.1)}  on ``\textit{airport}'', underscoring its strong trainability.

In summary, our contributions are threefold: (1) We propose FICGen, the first attempt to address the contextual illusion dilemma in degraded image generation. (2) We introduce a learnable dual-query mechanism to capture contextual frequency knowledge, which is injected into the latent diffusion space via a visual-frequency enhanced attention module, and further complemented by an instance coherence map and a spatial-frequency aggregation module to enable latent-space disentanglement and consolidation. (3) Extensive experiments on 5 degraded datasets show that our FICGen outperforms SOTA L2I methods in both quantitative and qualitative evaluations. Notably, FICGen can be seamlessly integrated as a plug-and-play component to enhance other diffusion-based generative models.
\section{Related Work}
\subsection{Text-Driven Image Synthesis}
Text-driven image synthesis aims to create visual content conditioned on diverse textual descriptions. Early endeavors focus on leveraging conditional GANs~\cite{reed2016generative, xu2018attngan, zhang2017stackgan, zhang2021cross} to generate realistic images. Recent advances in diffusion~\cite{ho2020denoising, song2020denoising, rombach2022high, nichol2021glide, ramesh2022hierarchical, saharia2022photorealistic, podell2023sdxl} and auto-regressive models~\cite{ramesh2021zero, gafni2022make, yu2022scaling} have revolutionized image synthesis by enabling stable training and producing high-fidelity images. Typically, DALL-E~\cite{ramesh2021zero} exemplifies an auto-regressive work that treats textual and visual tokens as a unified data sequence, demonstrating remarkable zero-shot capability. Among diffusion-based efforts, latent diffusion model (LDM)~\cite{rombach2022high} efficiently implements a denoising Markov chain in a low-dimensional latent space, striking a balance between generative quality and computational cost. Nevertheless, these text-prompted methods struggle with fine-grained control over object attributes (\textit{i.e., position}).
\subsection{Layout-Driven Image Synthesis}
Layout-driven image synthesis~\cite{zheng2023layoutdiffusion, chen2023geodiffusion, yang2023reco, mou2024t2i, li2023gligen} encodes geometric layouts as positional tokens to enable fine-grained controllability. However, these methods often suffer from attribute leakage, where certain layouts mistakenly attend to irrelevant regions. Multi-instance control~\cite{zhou2024migc, wang2024instancediffusion, wu2024ifadapter} addresses this by partitioning layouts into disjoint components, allowing each layout to focus on its corresponding region. Nonetheless, these approaches are confined to natural image domains.  AeroGen~\cite{tang2024aerogen} pioneers layout-conditioned generation in the context of remote sensing imagery, but it still struggles with semantic ambiguity and limited layout controllability. CC-Diff~\cite{zhang2024cc} further improves contextual coherence by modeling foreground–background relations, but overlooks frequency-aware cues and remains vulnerable to contextual illusion in various degraded scenes. In contrast, we propose a frequency-inspired contextual disentanglement paradigm, which introduces contextual frequency guidance into the generation process, delivering more realistic degraded images across diverse circumstances.
\label{sec:related}

\section{Method}
\begin{figure*}[!t]
\centering
\includegraphics[width=1.00\textwidth]{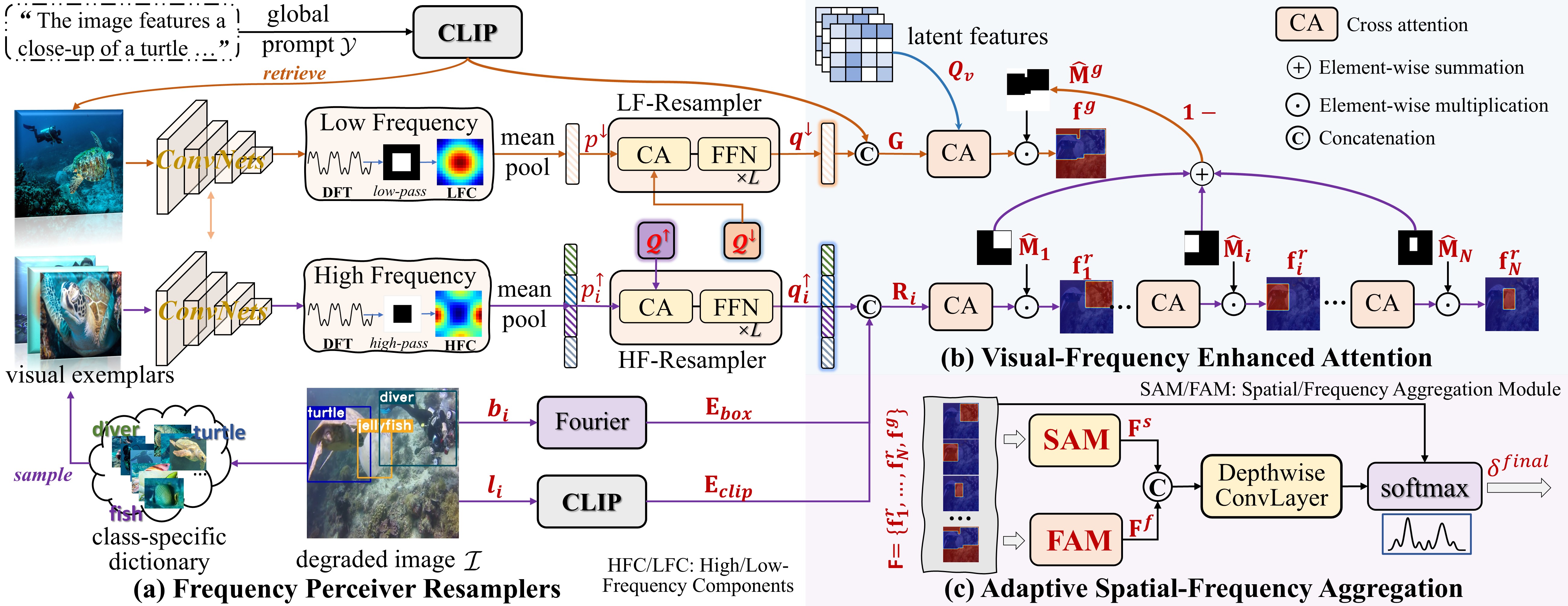}
\caption{\textbf{The overview pipeline of our FICGen}, consisting of (a) Frequency Perceiver Resamplers, (b) Visual-Frequency Enhanced Attention, and (c) Adaptive Spatial-Frequency Aggregation.}
\label{FIG:2}
\end{figure*}
\subsection{Preliminaries}\label{sec1}
\textbf{Latent Diffusion Model.} Unlike vanilla DDPM~\cite{ho2020denoising}, which generates images from Gaussian noise in pixel space, LDM~\cite{rombach2022high} operates in a compressed latent space. Specifically, an input image $\mathcal{I}$ is first encoded into a latent representation $z_{0}$, which is then diffused into a noisy latent state via the forward process: $z_{t}=\alpha_{t}z_{0}+\sigma_{t}\epsilon$, where $\alpha_{t}$ and $\sigma_{t}$ are the fixed schedule, $\epsilon \sim \mathcal{N}(0,1)$ is added standard Gaussian noise term and $t\in [0,T]$ denotes the timestep. Then, a denoising U-Net ($\epsilon_{\theta}$)~\cite{ronneberger2015u} is trained to predict the step-wise noise $\epsilon$, with the training objective $\mathcal{L}_{LDM}$ defined in Eq.\ref{eq1}.
\begin{equation}
\label{eq1}\mathop{min}\limits_{\theta}\mathcal{L}_{LDM}=\mathbb{E}_{z_{0},\epsilon\sim \mathcal{N}(0,1),t,\mathcal{\tau}}[||\epsilon-\epsilon_{\theta}(z_{t},t,\mathcal{\tau})||_{2}^{2}],
\end{equation}
where $\tau$ is the guidance embedding (\ie, text and bboxes).
\\
\textbf{Problem Formulation.} Given instance layouts comprising $N$ object bboxes: $\mathcal{B}=(\{l_{i},b_{i}\}_{i=1}^{N})$, where the location of $i$-th object $b_{i}=[x_{i}^{1},y_{i}^{1},x_{i}^{2},y_{i}^{2}]$ (\textit{i.e.}, top-left and bottom-right corners), and $l_{i}$ is its corresponding semantic category. $\mathcal{Y}$ denotes the global textual description of the degraded image. In this regard, our FICGen ($\mathcal{G}(\cdot)$) strives to generate a degraded image conditioned on both spatial layout and semantics: $\mathcal{I}=\mathcal{G}(\mathcal{B},\epsilon,\mathcal{Y})$.

\subsection{Bridge Queries with Frequency Prototypes}\label{sec2}
Existing L2I methods~\cite{chen2023geodiffusion, li2023gligen, zheng2023layoutdiffusion, gu2024roictrl, zhou2024migc} primarily emphasize foreground layout conditions to render instances in natural domains, where fronts and contexts share a relatively uniform frequency distribution. Nonetheless, the naive strategy is inadequate for degraded domains with contextual illusion. 

Exploring degraded contexts reveals that frequency disentanglement, a longstanding yet insightful concept, effectively mines weakened HF instances from the dominant LF surroundings. To this end, we seek to extract contextual frequency knowledge and transfer it into the latent space, thus achieving more realistic degraded generation via frequency-aware guidance. To achieve this, we face two main issues: 1) \textit{how to \textbf{represent} the frequency knowledge for degraded contexts?} and 2) \textit{how to \textbf{transfer} frequency knowledge into the latent diffusion space?} The answers are detailed below.

\textbf{Frequency Prototypes Representation.} 
For the former issue, our goal is to represent contextual frequency knowledge (\ie, textures) using frequency prototypes. To this end, we take two steps to manifest them, as depicted in Fig.~\ref{FIG:2}(a). Taking the foreground exemplar as an example, we first build a class-specific dictionary by collecting same-class instances from the training set. A degraded instance is then sampled and assigned to the $i$-th bbox based on its position. All selected instances form a foreground exemplar that is often overwhelmed in degraded scenarios. Second, stacked convolutional networks extract its intermediate feature maps $\mathbf{X} \in \mathbb{R}^{H \times W}$, which are then transformed into the frequency domain $\mathcal{F}(\mathbf{X})$ via a Discrete Fourier Transform (DFT):
\begin{equation}
\label{eq2}
    \textbf{X}_{\mathcal{F}}(u,v)=\frac{1}{H\times W}\sum_{h=0}^{H-1}\sum_{w=0}^{W-1}\textbf{X}(h,w)e^{-j2\pi(uh+vw)},
\end{equation}
where $\textbf{X}_{\mathcal{F}}$ is the complex-valued DFT output, $|v|$ and $|u|$ denote the width and height of normalised frequencies. Next, to isolate HF components, a binary mask $\textbf{M}_{\mathcal{F}}$ is applied:
\begin{equation}
\label{eq3}
\mathbf{M}_{\mathcal{F}}(u,v) = 
\begin{cases}
    1, & \text{if } u \notin \left[\frac{H}{2} - \gamma H,\ \frac{H}{2} + \gamma H\right], \\ 
       & \quad\ \ v \notin \left[\frac{W}{2} - \gamma W,\ \frac{W}{2} + \gamma W\right], \\
    0, & otherwise
\end{cases}
\end{equation}
where $\gamma$ controls the size of the HF region. The masked frequencies $\textbf{X}_{\mathcal{F}}\textbf{M}_{\mathcal{F}}$ are then enhanced using learnable channel-wise weights $\mathbf{W}_{\mathcal{F}}$, and projected back to the spatial domain:
\begin{equation}
\label{eq4}
    \textbf{X}^{\uparrow}=\mathcal{F}^{-1}(\textbf{X}_{\mathcal{F}}\textbf{M}_{\mathcal{F}})\cdot\mathbf{W}_{\mathcal{F}},
\end{equation}
where $\mathcal{F}^{-1}$ denotes the inverse DFT. The result $\textbf{X}^{\uparrow}$ represents the enhanced HF feature map. A similar process is applied to the complementary region using $1-\textbf{M}_{F}$, yielding the LF feature map $\textbf{X}^{\downarrow}$. Inspired by the prototypical learning, we adopt their mean pooled results as HF and LF prototypes, \ie, $\textbf{p}^{\uparrow}=\{p_{i}^{\uparrow}\}_{i=1}^{N}\in\mathbb{R}^{N\times d}$ and $p^{\downarrow}\in\mathbb{R}^{d}$, where $\textbf{p}^{\uparrow}$ corresponds to the HF representations of degraded instances, $p^{\downarrow}$ reflects the LF characteristics of surrounding contexts, and $d$ denotes the feature dimension.

\textbf{Frequency Perceiver Resamplers.}  To address the latter issue, inspired by the Perceiver architecture~\cite{ye2023ip}, we adopt a learnable dual-query mechanism that employs separate frequency resamplers to bridge HF and LF prototypes, thereby capturing contextual frequency cues:
\begin{align}
\label{eq5}
    \textbf{q}_{i}^{\uparrow} &= \text{HF-Resampler}(\mathcal{\textbf{Q}}^{\uparrow},\phi_{k1}^{r}(p_{i}^{\uparrow}),\phi_{v1}^{r}(p_{i}^{\uparrow})),\\
    \textbf{q}^{\downarrow} &=  \text{LF-Resampler}(\mathcal{\textbf{Q}}^{\downarrow},\phi_{k1}^{g}(p^{\downarrow}),\phi_{v1}^{g}(p^{\downarrow})),
\end{align}
where $\phi_{*}^{r},\phi_{*}^{g}\in\mathbb{R}^{d\times d}$ are linear projection layers. The dual queries $\mathcal{\textbf{Q}^{\uparrow}}$ and $\mathcal{\textbf{Q}^{\downarrow}}$ are learned through their respective resamplers, each composed of multiple transformer blocks~\cite{vaswani2017attention}, enabling continuous interactions between the dual queries and frequency prototypes, yielding frequency-aware instance and context tokens, \ie, $\textbf{q}_{i}^{\uparrow}$ and $\textbf{q}^{\downarrow}$. 

With the above dedicated design, the synergy between frequency prototypes and the dual-query mechanism enables the simultaneous perception of essential frequency cues and contextual appearances, which are then incorporated into the degraded image generation process.

\subsection{Contextual Frequency Knowledge Transfer}\label{sec3}
Previous L2I efforts~\cite{cheng2023layoutdiffuse, li2023gligen, chen2023geodiffusion, zhou2024migc} only focus on instance generation conditioned on layouts, while neglecting the surrounding context. This limitation is particularly detrimental in degraded scenarios involving contextual entanglement.

\textbf{Visual-Frequency Enhanced Attention.} To transfer frequency properties into the degraded generation, we employ a visual-frequency enhanced attention module (Fig.\ref{FIG:2}(c)) that injects frequency-aware tokens $\textbf{q}_{i}^{\uparrow}$ and $\textbf{q}^{\downarrow}$ into the latent diffusion space. Specifically, each bbox $b_i$ is first encoded using a Fourier embedding, followed by a multi-layer perceptron ($\textbf{E}_{box}$) to produce positional tokens. In parallel, its semantic class $l_i$ is encoded via a CLIP text encoder ($\textbf{E}_{clip}$)~\cite{radford2021learning}. Notably, $\textbf{q}_{i}^{\uparrow}$ is then integrated with them to provide frequency-aware cues for instance generation. As for the surrounding, $\textbf{q}^{\downarrow}$ is fused with text tokens extracted from the global caption $\mathcal{Y}$, thereby facilitating holistic contextual guidance.
\begin{align}
\label{eq6}
    \textbf{R}_{i}&=[\textbf{q}_{i}^{\uparrow};\textbf{E}_{clip}(l_{i});\textbf{E}_{box}\text{(Fourier($b_{i}$))}],\\
    \textbf{G}&=[\textbf{q}^{\downarrow};\textbf{E}_{clip}(\mathcal{Y})],
\end{align}
where $[;]$ represents concatenate operation.

It is preferable for each layout to affect only its corresponding local region. To this end, an instance coherence map $\hat{\textbf{M}}_{i}$ is utilized to isolate individual instances and enforce precise interactions between layout conditions and their associated latent features.
\begin{align}
\label{eq7}
    \mathbf{f}_{i}^{r}&=\text{softmax}(\frac{Q_{v}\phi_{k2}^{r}(\textbf{R}_{i})}{\sqrt{d}})\cdot\phi_{v2}^{r}(\textbf{R}_{i})\odot\hat{\textbf{M}}_{i},\\
    \mathbf{f}^{g}&=\text{softmax}(\frac{Q_{v}\phi_{k2}^{g}(\textbf{G})}{\sqrt{d}})\cdot\phi_{v2}^{g}(\textbf{G})\odot\hat{\textbf{M}^{g}},
\end{align} 
where $Q_{v}$ denotes the linear projection of latent features, $\odot$ is element-wise multiplication, and $\phi_{*}^{r}, \phi_{*}^{g}$ are linear layers. Finally, $\textbf{f}_{i}^{r}$ indicates the latent feature of the $i$-th instance, while $\textbf{f}^{g}$ corresponds to the contextual representation.

\textbf{Instance Coherence Map.} To avert feature contamination and attribute leakage, we first construct an instance-level mask $\hat{\mathbf{M}}_{i}$ from each bbox layout, as in ~\cite{zhou2024migc, wu2024ifadapter}, which enables precise generation of instance-specific latent features:
\begin{equation}
\label{eq8}
\hat{\mathbf{M}}_{i}(x,y) = 
\begin{cases}
    1, & \text{if } [x,y] \in b_{i}, \\
    0, & \text{if}  [x,y] \notin b_{i},
\end{cases}
\end{equation}
where $x$ and $y$ denote the mask coordinates. On the basis, we further separate foreground instances from their surroundings by constructing the complementary region $\hat{\mathbf{M}}^{g}$, which encompasses all areas outside the instance-level masks, thus enabling contextual disentanglement:
\begin{equation}
\label{eq9}
\hat{\mathbf{M}}^{g}=1-\sum_{i=1}^{N}\hat{\mathbf{M}}_{i}.
\end{equation}
Then, we combine $\textbf{f}_{i}^{r}$ and $\textbf{f}^{g}$ to produce the unified latent degraded representations $\textbf{F}=[\{\textbf{f}_{i}^{r}\}_{i=1}^{N},\textbf{f}^{g}]$, which serve as the key for degraded image generation.

\subsection{Adaptive Spatial-Frequency Aggregation}\label{sec4}
After harvesting the degraded representations $\textbf{F}$, we move forward to jointly consolidate and reconstruct them. In contrast to existing methods that either directly sum multiple instances~\cite{zhang2024cc} or combine them solely within the spatial domain~\cite{zhou2024migc}, which often suffer from ``object omission and merging'' for overlapping instances, as illustrated in Fig.~\ref{FIG:2}(c), we aim to adaptively aggregate the degraded instances and their surroundings in a spatial-frequency mixed manner. ``Spatial'' captures contextual relational dependencies, while ``frequency'' emphasizes fine-grained attributes across instances, \ie, boundary sharpness and textures, which can be formulated as follows:
\begin{equation}
 \textbf{F}^{s}=\textbf{SAM}([\sum_{i=1}^{N}\textbf{f}_{i}^{r},\textbf{f}^{g}]),\textbf{F}^{f}=\textbf{FAM}([\sum_{i=1}^{N}\textbf{f}_{i}^{r},\textbf{f}^{g}]), \\
\end{equation}
\begin{equation}
    \delta^{final}=\sum_{i=1}^{N+1}\text{softmax}(\delta^{fusion})[\sum_{i=1}^{N}\textbf{f}_{i}^{r},\textbf{f}^{g}],
\end{equation}
where $\textbf{SAM}$ and $\textbf{FAM}$ represent the spatial and frequency aggregation modules, respectively, consisting of standard self- and frequency-attention mechanisms. $\delta^{fusion}=\zeta*([\textbf{F}^{s},\textbf{F}^{f}])$, where $\zeta$ is is a learnable single-layer depthwise convolution to enhance local perception. Finally, the refined degraded representation $\delta^{final} \in \mathbb{R}^{h \times w \times c}$ is decoded into the target degraded image.
\subsection{Learning Procedure}\label{sec5}
During training, the parameters of the pretrained LDM remain fixed, with optimization focused solely on FICGen. The learning objective integrates the original mean-square error loss and the contextual frequency guidance:
\begin{equation}
\label{eq10}\mathop{min}\limits_{\theta^{'}}\mathcal{L}_{FICGen}=\mathbb{E}_{z_{0},\epsilon\sim \mathcal{N}(0,1),t,\mathcal{Y}}[||\epsilon-\mathcal{G}_{\theta,\theta^{'}}(z_{t},t,\mathcal{Y},\mathcal{B},\mathcal{Q})||_{2}^{2}],
\end{equation}
where $\mathcal{Q}=[\{\textbf{q}_{i}^{\uparrow}\}_{i=1}^{N},\textbf{q}^{\downarrow}]$ denotes the set of frequency-aware tokens. While $\theta$ and $\theta^{'}$ denote the frozen and trainable parameters of the pretrained LDM and our FICGen, respectively.

\section{Experiment}
\subsection{Experimental Setups}
\textbf{Datasets.} Our FICGen is evaluated on five degraded-context datasets: ExDARK~\cite{loh2019getting}, RUOD~\cite{fu2023rethinking}, DAWN~\cite{kenk2020dawn}, DIOR~\cite{li2020object}, and blurred VOC 2012~\cite{everingham2010pascal}. More details are provided in the \textbf{\textit{Supplementary Material}}.

\begin{table}
\centering
\scalebox{0.75}{
\begin{tabular}{ccc|ccc}
\toprule[1.0pt]%
Detector&  Method &FID $\downarrow$ & mAP $\uparrow$ & AP\_{50} $\uparrow$ &AP\_{75} $\uparrow$ \\
\hline \rowcolor{gray!5}
\multicolumn{6}{c}{\textcolor{red}{\textbf{DIOR-H}} \cite{li2020object}} \\
Faster R-CNN&  Oracle&-&33.4&55.6&35.0\\
\hline
Faster R-CNN & MIGC~\cite{zhou2024migc}&31.64&21.8&38.4&17.5\\
Faster R-CNN & CC-Diff~\cite{zhang2024cc}&\textcolor{red}{30.88}&\underline{23.6}&\underline{42.4}&\underline{21.4}\\
\rowcolor{gray!10}
Faster R-CNN & FICGen (ours)&\underline{31.25}&\textcolor{red}{27.6}&\textcolor{red}{48.7}&\textcolor{red}{27.6}\\
\hline \rowcolor{gray!5}
\multicolumn{6}{c}{\textcolor{purple}{\textbf{RUOD}} \cite{fu2023rethinking}}\\
Faster R-CNN & Oracle&-&50.5&80.2&54.4\\
\hline
Faster R-CNN & MIGC~\cite{zhou2024migc}&26.50&27.2&54.1&24.6\\
Faster R-CNN&CC-Diff~\cite{zhang2024cc}&\underline{25.21}&\underline{29.7}&\underline{58.4}&\underline{27.9}\\
\rowcolor{gray!10}
Faster R-CNN&FICGen (ours)&\textcolor{red}{25.10}&\textcolor{red}{37.0}&\textcolor{red}{68.6}&\textcolor{red}{36.5}\\
\hline \rowcolor{gray!5}
\multicolumn{6}{c}{\textcolor{orange}{\textbf{blurred VOC 2012}} \cite{everingham2010pascal}}\\
Faster R-CNN & Oracle&-&31.5&56.5&32.4\\
\hline
Faster R-CNN & MIGC~\cite{zhou2024migc}&62.66&34.7&65.8&33.1\\
Faster R-CNN&CC-Diff~\cite{zhang2024cc}&\underline{62.20}&\underline{36.7}&\underline{67.6}&\underline{36.3}\\
\rowcolor{gray!10}
Faster R-CNN&FICGen (ours)&\textcolor{red}{58.02}&\textcolor{red}{40.7}&\textcolor{red}{70.3}&\textcolor{red}{42.7}\\
\hline \rowcolor{gray!5}
\multicolumn{6}{c}{\textcolor{blue}{\textbf{ExDARK}} \cite{loh2019getting}}\\
Cascade R-CNN & Oracle&-&37.2&65.8&37.8\\
\hline
Cascade R-CNN & MIGC~\cite{zhou2024migc}&45.76&32.4&63.5&29.5\\
Cascade R-CNN& CC-Diff~\cite{zhang2024cc}&\underline{44.26}&\underline{35.1}&\underline{65.6}&\underline{34.1}\\
\rowcolor{gray!10}
Cascade R-CNN & FICGen (ours)&\textcolor{red}{42.40}&\textcolor{red}{42.5}&\textcolor{red}{73.0}&\textcolor{red}{45.1}\\
\hline \rowcolor{gray!5}
\multicolumn{6}{c}{\textcolor{green}{\textbf{DAWN}} \cite{kenk2020dawn}}\\
Cascade R-CNN& Oracle&-&27.3&46.4&26.3\\
\hline
Cascade R-CNN & MIGC~\cite{zhou2024migc}&70.10&12.6&32.3&8.6 \\            
Cascade R-CNN & CC-Diff~\cite{zhang2024cc}&\underline{68.56}&\underline{15.7}&\underline{33.9}&\underline{14.8}  
\\
\rowcolor{gray!10}
Cascade R-CNN & FICGen (ours)&\textcolor{red}{68.31}&\textcolor{red}{22.6}&\textcolor{red}{44.3}&\textcolor{red}{21.5}\\
\bottomrule[1.0pt]
\end{tabular}
}
\caption{Quantitative comparison of fidelity and alignment on various degraded conditions. The performance is evaluated by off-the-shelf detectors Faster R-CNN~\cite{ren2015faster} and Cascade R-CNN~\cite{cai2018cascade} on synthetic test set. ``\textit{Oracle}'' denotes the real test baseline. The best results are in \textcolor{red}{red} and the second best results are in \underline{underline}.}
\label{table1}
\end{table}
\textbf{Implementation Details}. Our FICGen is built upon the pre-trained SDv1.5 checkpoint and deployed only on the mid-level (\ie, 8$\times$8) and lowest-resolution (\ie, 16$\times$16) decoder layers. For training, all images are resized to 512$\times$512. We adopt the AdamW~\cite{loshchilov2017decoupled} optimizer with a learning rate of 1e$^{-4}$, $\beta_{1}=0.9$, and $\beta_{2}=0.999$, and fine-tune for 300 epochs on 8 A100 with a batch size of 320. For inference, we use the EulerDiscreteScheduler~\cite{karras2022elucidating} with 50 sampling steps and a classifier-free guidance scale~\cite{ho2022classifier} of 7.5. \\
\textbf{Evaluation Metrics.} Following ~\cite{chen2023geodiffusion,li2023gligen}, we comprehensively evaluate our FICGen from three aspects: \textbf{fidelity}, \textbf{alignment}, and \textbf{trainability}. Fidelity refers to the perceptual quality between synthetic and real images, measured by the Fréchet Inception Distance (FID)~\cite{heusel2017gans}. While alignment assesses spatial and semantic correspondences between layout conditions and generated instances, quantified via COCO-style average precision~\cite{lin2014microsoft} with off-the-shelf detectors. Trainability examines the potential of synthetic images for auxiliary downstream training, assessed by mean Average Precision (mAP) under various training settings.

\begin{figure*}[!t]
\centering
\includegraphics[width=1.00\textwidth]{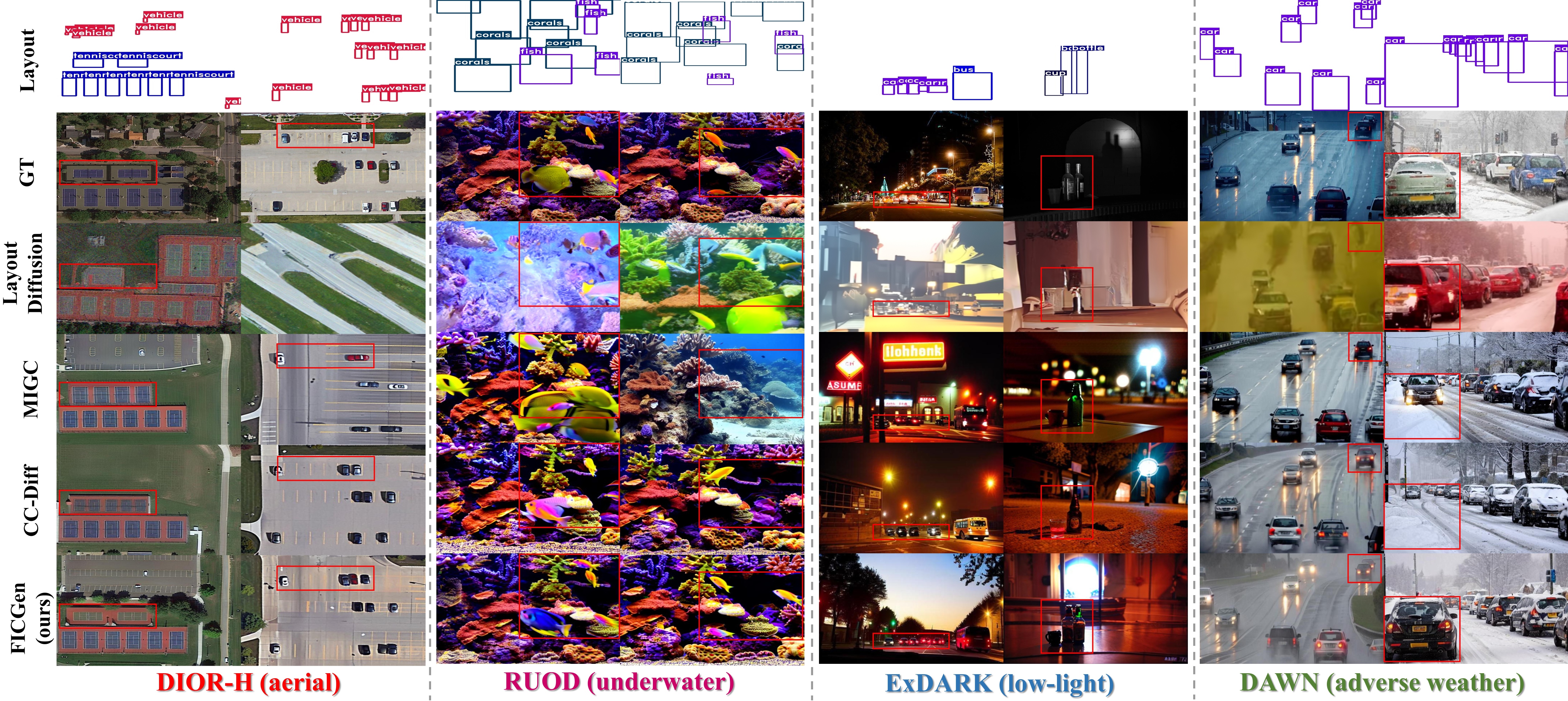}
\caption{Qualitative comparison of degraded images generated by different L2I methods across various scenarios. Zoom in for better details.}
\label{FIG:3}
\end{figure*}

\begin{table}
\centering
\resizebox{1\columnwidth}{!}{
\setlength{\tabcolsep}{3.5mm}
\renewcommand{\arraystretch}{1.0}
\begin{tabular}{cc|ccc}
\toprule[1.0pt]%
\multirow{2}*{Method} & \multirow{2}*{FID $\downarrow$}&  \multicolumn{3}{c}{YOLO Score $\uparrow$} \\
\cmidrule{3-5}%
& &mAP&AP\_{50}&AP\_{75} \\
\hline
Oracle&-&39.2&57.1&43.0\\
\hline
LostGAN~\cite{sun2019image} & 57.10 & 14.3 & 27.3 & 15.2\\
LayoutDiffusion~\cite{zheng2023layoutdiffusion}&45.31&20.0&37.4&19.3\\
ReCo~\cite{yang2023reco} & 42.56 & 21.1 & 40.7 & 23.1\\
MIGC~\cite{zhou2024migc} & 31.64 & 24.5 & 41.8 & 25.6\\
GLIGEN~\cite{li2023gligen} & 41.31 & 25.8 & 44.4 & 27.8\\
CC-Diff~\cite{zhang2024cc} & \textcolor{red}{30.88} & 26.4 & 44.2 & 28.5\\
AeroGen~\cite{tang2024aerogen} & 38.57 & \underline{29.8} & \textcolor{red}{54.2} & \underline{31.6}\\
\rowcolor{gray!10}
FICGen (ours) & \underline{31.25} & \textcolor{red}{31.2} & \underline{49.9} & \textcolor{red}{34.6}\\
\bottomrule[1.0pt]
\end{tabular}
}
\caption{Quantitative comparison on the DIOR-H \textit{test} set, with detection performance evaluated using a pre-trained YOLOv8~\cite{YOLOv8}.}
\label{table2}
\end{table}

\subsection{Main Results}
\subsubsection{Quantitative Evaluations}
\textbf{Fidelity and Alignment.} Tab.~\ref{table1} reports the quantitative comparison of fidelity and alignment via reporting the COCO-style AP, where pre-trained Faster R-CNN (R50)~\cite{ren2015faster} and Cascade R-CNN (R50)~\cite{cai2018cascade} are employed to infer synthetic images, with the results evaluated against corresponding real annotations. Notably, our FICGen successfully outperforms the existing MIGC~\cite{zhou2024migc} and CC-Diff~\cite{zhang2024cc} in terms of both FID and AP metrics, achieving remarkable FID scores of 25.10 and 42.40 on the RUOD and ExDARK, respectively. Moreover, for remote sensing object detection, FICGen reaches 82\% of the mAP relative to real data (\ie, 27.6 \textit{vs.} 33.4), substantially outperforming MIGC (21.8) and CC-Diff (23.6) by a large margin. Intriguingly, FICGen even exceeds the real ExDARK test set in AP, highlighting its ability to generate degraded instances that precisely adhere to layouts while bridging the real-to-synthetic domain gap. 

\begin{table}
\centering
\setlength{\tabcolsep}{2.0mm}
\renewcommand{\arraystretch}{1}
\scalebox{0.75}{
\begin{tabular}{c|ccc|ccc}
\toprule[1.0pt]%
\multirow{2}*{Method} & \multicolumn{3}{c}{ExDARK~\cite{loh2019getting}}&\multicolumn{3}{|c}{RUOD~\cite{fu2023rethinking}} \\
\cmidrule{2-7}%
&mAP$\uparrow$&AP\_{50}$\uparrow$&AP\_{75}$\uparrow$&mAP$\uparrow$&AP\_{50}$\uparrow$&AP\_{75}$\uparrow$\\
\hline
Oracle & 38.3 & 67.8 & 39.8 & 58.4&85.5&64.8\\
\hline
MIGC~\cite{zhou2024migc} & 28.6 & 59.2 & 24.1 & 21.1&43.9&17.9\\
CC-Diff~\cite{zhang2024cc} & \underline{31.3} & \underline{61.8} & \underline{28.8} & \underline{29.7}&\underline{57.8}&\underline{28.0}\\
\rowcolor{gray!10}
FICGen (ours) & \textcolor{red}{38.5} & \textcolor{red}{68.5} & \textcolor{red}{39.5} & \textcolor{red}{37.1}&\textcolor{red}{67.1}&\textcolor{red}{36.7}\\
\bottomrule[1.0pt]
\end{tabular}
}
\caption{Quantitative comparison on ExDARK and RUOD \textit{test} set, evaluated by a pre-trained Deformable-DETR~\cite{zhu2020deformable}.}
\label{table3}
\end{table}

\begin{table*}
\centering
\resizebox{1\textwidth}{!}{
\begin{tabular}{cc|cc|cccccccc|ccc}
\toprule[1.0pt]%
\multirow{2}*{Detector}& \multirow{2}*{Backbone}  & \multicolumn{2}{|c|}{Training Set} & \multicolumn{8}{|c|}{Object Detection (AP) for Sampled Classes /\%} &  \multirow{2}*{mAP}&\multirow{2}*{AP\_{50}}&\multirow{2}*{AP\_{75}} \\
\cmidrule{3-4}%
\cmidrule{5-12}%
 & &\# \textbf{\textit{Real}} &\# \textbf{\textit{Synthetic}} &bicycle &bus &motorbike &boat &car & cup & bottle &dog \\
\hline \rowcolor{gray!16}
\multicolumn{15}{c}{\textit{Train with Full Real Data}} \\
RetinaNet & PVT-M & R:5.1\textit{k} & - &44.8 &58.2 & 31.3 &  31.1 & 36.0 & 25.5 & 28.7& 47.8 &35.2 &65.4 & 33.9 \\
Faster R-CNN & X101 & R:5.1\textit{k} & - &49.0 &60.8 & 35.6 &  30.1 & 39.5 & 30.9 & 34.1& 49.3 &38.6 &69.3 & 39.6 \\
\hline \rowcolor{gray!16}
\multicolumn{15}{c}{\textit{{Train with Pure Synthetic Data}}} \\
RetinaNet & PVT-M &- & MIGC (S: 5.1\textit{k}) &32.8 &46.7 & 20.4 &  19.9 & 22.5 & 14.1 & 16.0& 33.4 &23.5 &49.0 & 19.1 \\
Faster R-CNN & X101 & - & MIGC (S: 5.1\textit{k}) &33.6 &48.2 & 18.5 &  20.4 & 23.1 & 17.0 & 21.1& 33.2 &24.7 &49.4 & 22.1 \\
RetinaNet & PVT-M &- & CC-Diff (S: 5.1\textit{k}) &34.3 &45.5 & 21.6 &  20.2 & 22.0 & 15.4 & 16.6& 34.9 &24.1 &49.9 & 20.6 \\
Faster R-CNN & X101 & - & CC-Diff (S: 5.1\textit{k}) &38.0 &47.5 & 21.5 &  19.1 & 24.3 & 18.3 & 21.8& 33.8 &25.5 &52.3 & 21.7\\
\rowcolor{gray!10}
RetinaNet & PVT-M & - & FICGen (S: 5.1\textit{k}) &35.1 &50.4 & 23.2 &  20.0 & 24.9 & 16.2 & 16.8 & 33.3 &25.3 &51.0& 22.1 \\
\rowcolor{gray!10}
Faster R-CNN& X101 & - & FICGen (S: 5.1\textit{k}) &38.1 &49.9 & 21.7 &  22.0 & 24.3 & 18.9 & 20.4 & 36.4 &26.7 &51.4 & 25.1 \\
\hline \rowcolor{gray!16}
\multicolumn{15}{c}{\textit{{Train with Synthetic Data and Finetune on Real Data}}} \\
RetinaNet & PVT-M & R:5.1\textit{k} & MIGC (S: 5.1\textit{k}) &47.3 &62.1 & 34.7 &  32.6 & 37.4 & 27.6 & 29.6 & 49.0 &37.1 &67.5 & 36.9 \\
Faster R-CNN & X101 & R:5.1\textit{k} & MIGC (S: 5.1\textit{k}) &51.3 &61.8 & 38.0 & 31.5 & 40.0 & 32.1& \underline{35.8} &49.7 &39.9 & \underline{70.7}&40.9 \\
RetinaNet & PVT-M & R:5.1\textit{k} & CC-Diff (S: 5.1\textit{k}) &47.9 &60.9&34.3 & 31.5 &  36.4 & 27.1 & 30.5 & 48.1 & 37.0 &67.0 &35.8 \\
Faster R-CNN & X101 & R:5.1\textit{k} & CC-Diff (S: 5.1\textit{k}) &50.2 &62.2 & \textcolor{red}{38.7} & 31.1 & \textcolor{red}{40.6} & 32.0& 35.4 &49.4 &39.8 & 70.4&40.7 \\
\rowcolor{gray!10}
RetinaNet & PVT-M & R:5.1\textit{k} & FICGen (S: 5.1\textit{k}) &47.5 &62.2 & 35.7 &  \textcolor{red}{33.7} & 37.8 & 27.6 & 30.9 & 47.6 &37.6 &67.7 & 36.8 \\
\rowcolor{gray!10}
Faster R-CNN & X101 & R:5.1\textit{k} & FICGen(S: 5.1\textit{k}) &\underline{51.4} &\underline{62.9} & 38.2 & \underline{32.8} & 40.3 & \textcolor{red}{33.1}& 35.6 &\underline{49.9} &\textcolor{red}{40.6} & \textcolor{red}{70.9}&\textcolor{red}{42.1} \\

\hline \rowcolor{gray!16}
\multicolumn{15}{c}{\textit{{Train with Real \& Synthetic Data}}} \\
Faster R-CNN & X101 & R:5.1\textit{k} & MIGC (S: 5.1\textit{k}) &\textcolor{red}{51.7} &\textcolor{red}{63.0} & 36.2 &  31.7 & 39.7 & 31.8 & 35.2& 49.6 &39.5 &69.7 & 39.8 \\
Faster R-CNN & X101 & R:5.1\textit{k} & CC-Diff (S: 5.1\textit{k}) &50.9 &61.6 & 36.5 &  31.8 & 40.2 & 31.6 & 34.7& \textcolor{red}{50.3} &39.5 &70.0 & 40.7 \\
\rowcolor{gray!10}
Faster R-CNN & X101 & R:5.1\textit{k} & FICGen (S: 5.1\textit{k}) &50.5 &61.8 & \underline{38.6} &  31.6 & \underline{40.4} & \underline{32.6} & \textcolor{red}{36.5}& 49.1 &\underline{40.1} &70.5 & \underline{41.9} \\
\bottomrule[1.0pt]
\end{tabular}
}
\caption{Quantitative comparison of trainability on the ExDARK dataset, evaluated by RetinaNet~\cite{lin2017focal} (PVT-M~\cite{wang2021pyramid}) and Faster R-CNN~\cite{ren2015faster} (X101~\cite{xie2017aggregated}) under three training settings.}\label{table4}
\end{table*}

\begin{table*}
\centering
\resizebox{1\textwidth}{!}{
\begin{tabular}{c|cccc|cccc|cccc|cccc}
\toprule[1.0pt]%
\multirow{2}*{Method}&  \multicolumn{4}{c|}{\textcolor{red}{DIOR-H} \cite{li2020object}}&\multicolumn{4}{c|}{\textcolor{green}{DAWN} \cite{kenk2020dawn}}&\multicolumn{4}{c|}{\textcolor{blue}{ExDARK} \cite{loh2019getting}}&\multicolumn{4}{c}{\textcolor{cyan}{VOC 2012} \cite{everingham2010pascal}}\\
\cmidrule{2-17}%
&\textit{airport} $\uparrow$&mAP $\uparrow$&AP\_{50} $\uparrow$&AP\_{75} $\uparrow$&\textit{bus} $\uparrow$&mAP $\uparrow$&AP\_{50} $\uparrow$&AP\_{75} $\uparrow$&\textit{motorbike} $\uparrow$&mAP $\uparrow$&AP\_{50} $\uparrow$&AP\_{75} $\uparrow$&\textit{boat} $\uparrow$&mAP $\uparrow$&AP\_{50} $\uparrow$&AP\_{75} $\uparrow$\\
\hline
Oracle&32.2&33.4&55.6&35.0&21.8&24.8&46.0&23.0&33.3&35.8&67.4&34.5&42.9&48.3&76.8&52.5\\
\hline
MIGC~\cite{zhou2024migc}&36.9&34.1&\underline{56.1}&36.1&21.6&24.9&43.3&24.6&\underline{34.4}&\underline{37.1}&\underline{67.8}&\underline{36.6}&42.5&49.4&77.2&54.2\\
CC-Diff~\cite{zhang2024cc}&\underline{37.6}&\underline{34.3}&\underline{56.1}&36.3&\underline{26.2}&\underline{25.8}&\textcolor{red}{46.1}&\underline{25.8}&34.1&37.0&67.7&\underline{36.6}&\underline{44.4}&\underline{49.6}&\underline{77.3}&\underline{54.3}\\
\rowcolor{gray!10}
FICGen (ours)&\textcolor{red}{38.1}&\textcolor{red}{35.5}&\textcolor{red}{56.8}&\textcolor{red}{37.6}&\textcolor{red}{27.5}&\textcolor{red}{25.9}&\underline{44.5}&\textcolor{red}{26.5}&\textcolor{red}{37.1}&\textcolor{red}{37.7}&\textcolor{red}{68.0}&\textcolor{red}{37.7}&\textcolor{red}{47.3}&\textcolor{red}{50.5} & \textcolor{red}{77.4} & \textcolor{red}{55.4}\\
\bottomrule[1.0pt]
\end{tabular}
}
\caption{Quantitative comparison of trainability on the DIOR, DAWN and ExDARK and natural VOC 2012 datasets, with detection evaluated using the Faster R-CNN (R50) object detector.}
\label{table5}
\end{table*}
To further verify the generative applicability across different detectors, we evaluate FICGen using YOLOv8~\cite{YOLOv8} and Deformable-DETR~\cite{zhu2020deformable}, as shown in Tabs.~\ref{table2} and \ref{table3}. FICGen surpasses AeroGen by 1.4 and 3.0 in mAP and AP$_{75}$ on the DIOR-H, and outperforms CC-Diff by 7.2 and 7.4 on the ExDARK and RUOD, respectively. 
%

\begin{figure}[!t]
\centering
\includegraphics[width=0.48\textwidth]{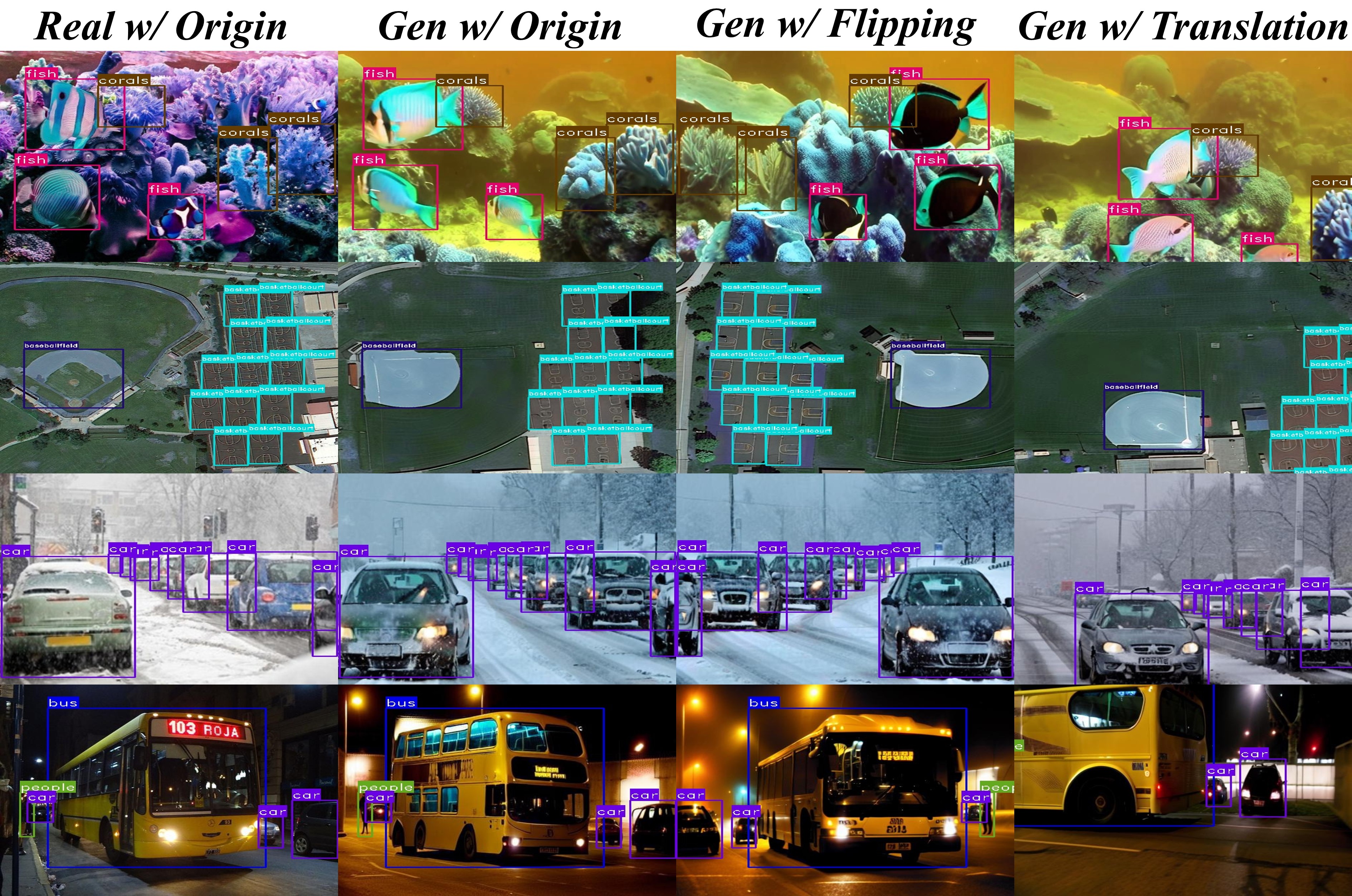}
\caption{Visualization results generated by our FICGen on unseen geometric layouts augmented via flipping and translation.}
\label{FIG:4}
\end{figure}

\textbf{Trainability.} Following the trainability protocol in~\cite{chen2023geodiffusion}, we use augmented annotations as layout inputs to generate synthetic training images and mix them with real ones, effectively doubling the training set. As shown in the last three rows of Tab.~\ref{table4}, we \textbf{demonstrate for the first time} that synthetic degraded images can significantly enhance the detection performance of downstream object detectors (\ie, \textbf{+2.4} for bottle and \textbf{+3.0} for motorbike). Moreover, we observe that training with synthetic images followed by fine-tuning on real ones yields the best performance. Tab.~\ref{table5} further provides a comparative analysis under various degraded conditions. Crucially, FICGen consistently achieves the highest gains, boosting AP by \textbf{5.7} for ``bus'' (DAWN) and \textbf{3.8} for ``motorbike'' (ExDARK), highlighting its efficacy in mitigating data scarcity in degraded scenarios.

\textbf{Generalization.} To evaluate the generalization of our FICGen, we conduct further experiments not only in severely degraded scenarios such as low-light and underwater conditions with horizontal layout control, but also in blurred and natural VOC 2012~\cite{everingham2010pascal}, and oriented DIOR-R~\cite{cheng2022anchor}. Extensive results demonstrate that synthetic images generated by FICGen exhibit superior quality and alignment. Further details are provided in the \textbf{\textit{Supplementary Material}}.
\subsubsection{Qualitative Evaluations}
\begin{figure}[!t]
\centering
\includegraphics[width=0.490\textwidth]{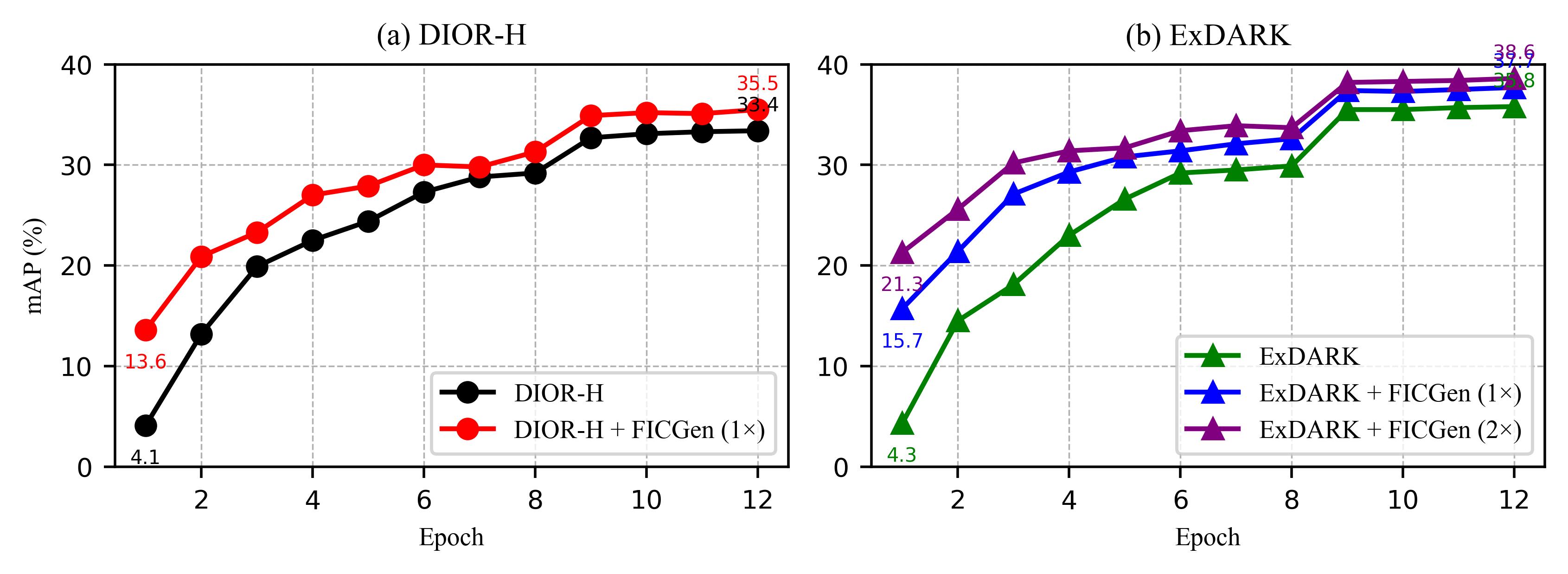}
\caption{Performance convergence curves when generated degraded images are incorporated into downstream detection.}
\label{FIG:5}
\end{figure}

\begin{figure}[!t]
\centering
\includegraphics[width=0.485\textwidth]{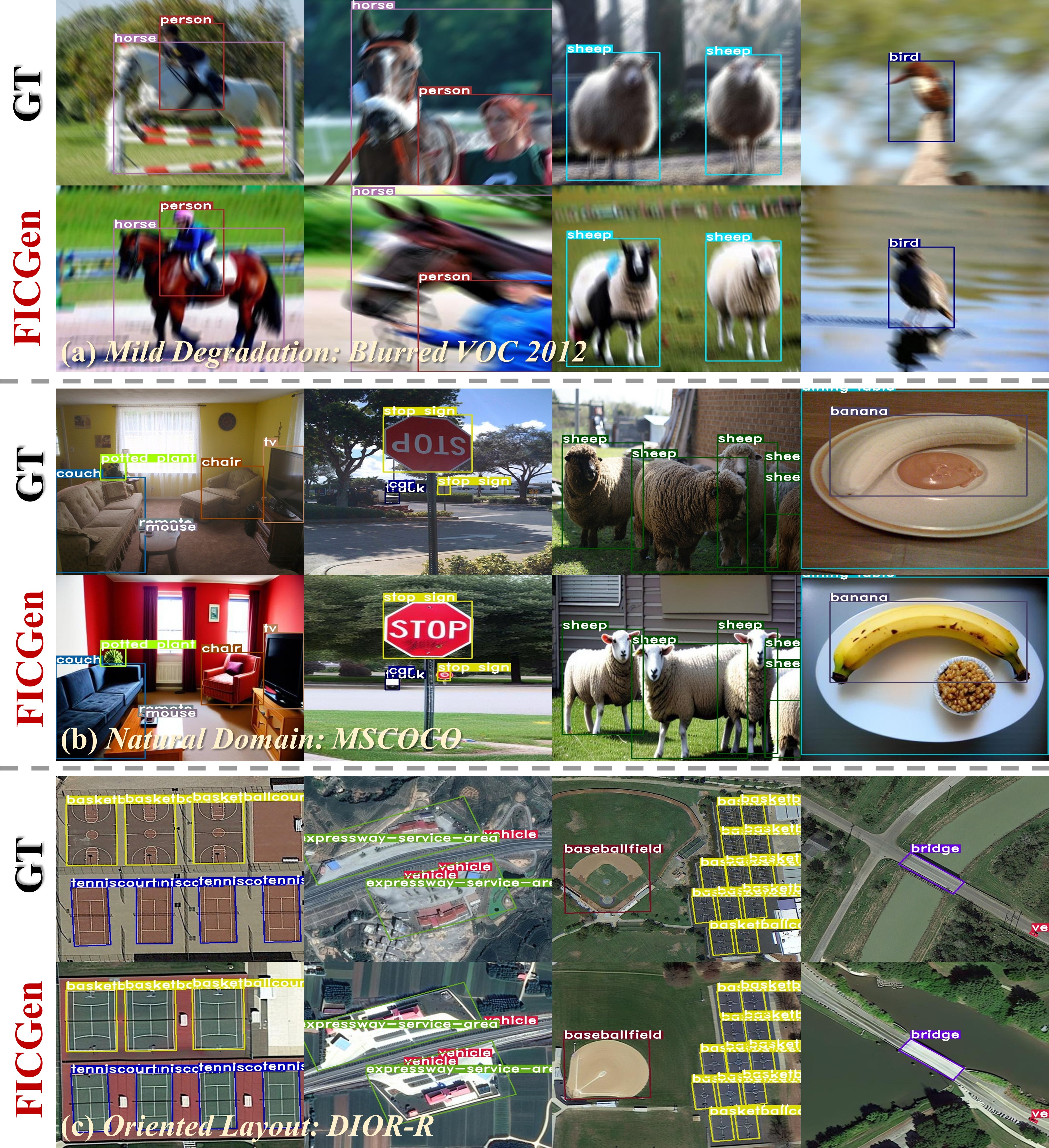}
\caption{Generalization results of FICGen on mildly blurred and natural scenes, as well as under oriented layout control.}
\label{FIG:7}
\end{figure}

Fig.~\ref{FIG:3} presents a visual comparison of degraded images generated by various L2I methods. Our FICGen demonstrates superior performance in two key aspects: \textbf{(a) Enhanced Layout Coherence.} It resolves the dilemma of ``object omission and merging'', accurately aligning generated instances with the provided layouts in terms of quantity, scale, and position (\ie, aerial tennis courts and rainy cars), and effectively handling small objects (\ie, aerial vehicles). \textbf{(b) Strong Occlusion Handling.} In heavily occluded scenarios (\ie, underwater corals and fish), FICGen reconstructs dense instances while maintaining spatial consistency, whereas MIGC and CC-Diff fail to enforce overlapped layout constraints. Furthermore, Fig.~\ref{FIG:4} shows that FICGen generalizes effectively to unseen layouts, including those extending beyond image boundaries. As illustrated in Fig.~\ref{FIG:5}, the synthesized degraded images generated by FICGen enhance downstream detection performance by improving mAP and accelerating training convergence. Finally, Fig.~\ref{FIG:7} highlights the generalization capability of FICGen in blurred and natural scenes, as well as under oriented layout controls. Additional visualizations are provided in the \textbf{\textit{Supplementary Material}}.

\begin{table}
\centering
\resizebox{1\columnwidth}{!}{
\setlength{\tabcolsep}{5.0mm}
\renewcommand{\arraystretch}{0.9}
\begin{tabular}{ccc|ccc}
\toprule[1.0pt]%
FPR&VFEA&ASFA & mAP &AP\_{50} &AP\_{75}\\
\hline
{\XSolidBrush}&{\XSolidBrush}&{\XSolidBrush}&15.7&34.8&12.0\\
{\XSolidBrush}&{\CheckmarkBold}&{\XSolidBrush}&23.2&44.6&22.7\\
{\CheckmarkBold}&{\XSolidBrush}&{\CheckmarkBold}&18.5&36.3&13.9\\
{\CheckmarkBold}&{\CheckmarkBold}&{\XSolidBrush}&25.5&46.3&25.1\\
{\XSolidBrush}&{\CheckmarkBold}&{\CheckmarkBold}&25.6&46.9&25.3\\
\hline
{\CheckmarkBold}&{\CheckmarkBold}&{\CheckmarkBold}&27.6&48.7&27.6 \\
\bottomrule[1.0pt]
\end{tabular}
}
\caption{Ablation study of FPR, VFEA, and ASFA modules, evaluated by the pre-trained Faster R-CNN (R50) on the DIOR test set.}
\label{table8}
\end{table}
\subsection{Ablation Studies}
\textbf{Frequency Perceiver Resamplers}. The foundation of frequency prototype learning lies in the FPR module, which combines a dual-query mechanism with contextual frequency knowledge. As shown in the last two rows of Tab.~\ref{table8}, FPR significantly improves the alignment scores from 25.6/46.9/25.3 to 27.6/48.7/27.6, demonstrating the effectiveness of contextual frequency guidance.

\textbf{Visual-Frequency Enhanced Attention}. VFEA serves as a bridge to inject frequency knowledge (\ie, boundaries and textures) into the latent space while integrating the instance coherence map for contextual disentanglement. As shown in the second row of Tab.~\ref{table8}, even without FPR and ASFA, VFEA achieves competitive performance over the baseline, reaching 23.2/44.6/22.7 in AP.

\textbf{Adaptive Spatial-Frequency Aggregation}. ASFA is designed to reconstruct refined degraded representations by integrating spatial and frequency perspectives. As evidenced by the results in the fourth and sixth rows of Tab.~\ref{table8}, incorporating ASFA improves the average precision (AP) from 25.5/46.3/25.1 to 27.6/48.7/27.6. This substantial gain underscores its critical role in consolidating contextual information and enhancing the fidelity of degraded image generation.

\section{Conclusion and Future Work}
In this paper, we introduce the task of degraded image generation and define the contextual illusion dilemma as its core challenge. To address this, we propose a frequency-inspired generative paradigm, FICGen. We first extract frequency prototypes from degraded contexts and employ a learnable dual-query mechanism to mine contextual frequency knowledge, which is then transferred into the latent diffusion space via a visual-frequency enhanced attention, coupled with an instance coherence map and a spatial-frequency aggregation module, to disentangle foreground instances from their surroundings and consolidate them into refined degraded representations. Extensive experiments on 5 degraded image datasets validate the effectiveness of FICGen. In future work, we plan to extend this frequency-inspired paradigm to 3D content generation for broader applications. 
\label{sec:conclusion}

\section*{Acknowledgements}
This work is partially supported by grants from the National Natural Science Foundation of China (No. 62132002 and No. 62202010), Beijing Nova Program (No.20250484786), and the Fundamental Research Funds for the Central Universities.

{
    \small
    \bibliographystyle{ieeenat_fullname}
    \bibliography{main}
}

\clearpage
\appendix

\twocolumn[
\begin{center}
    {\LARGE \bf 
    FICGen: Frequency-Inspired Contextual Disentanglement for Layout-driven Degraded Image Generation
    \par}
    \vspace{1em}
    {\large Supplementary Material \par}
    \vspace{2em}
\end{center}
]
The supplementary material provides a more comprehensive evaluation of our proposed \textit{FICGen} for degraded image synthesis, which is organized as follows: 
\begin{itemize}
\item Section~\ref{sec1_}: Dataset Introductions.
\item Section~\ref{sec2_}: Implementation Details.
\item Section~\ref{sec3_}: Additional Experimental Results.
\item Section~\ref{sec4_}: Generative Visualizations.
\item Section~\ref{sec5_}: Limitations and Future Work.
\end{itemize}

\section{Dataset Introductions}
\label{sec1_}
To comprehensively evaluate the generative effectiveness of our \textit{FICGen} across diverse degraded conditions, we conduct extensive experiments on \textbf{five} widely used benchmarks: ExDARK~\cite{loh2019getting}, DIOR~\cite{li2020object}, RUOD~\cite{fu2023rethinking}, DAWN~\cite{kenk2020dawn}, and blurred VOC 2012~\cite{everingham2010pascal}. Tab.~\ref{table7} summarizes the key statistics of each dataset, including the numbers of training and testing samples, total annotated instances, and degradation types. Two considerations are noteworthy:

\textbf{(1)}. Due to the relatively small size of DAWN, we employ conventional data augmentation techniques—such as scaling, translation, and horizontal flipping—to expand the training set from 590 to 5,544 samples. Additionally, given the extreme scarcity of ``bicycle'' instances in DAWN, this category is omitted from the evaluation.

\textbf{(2)}. Following~\cite{sayed2021improved}, we generate blur kernels on the fly to construct the blurred VOC 2012~\cite{everingham2010pascal} dataset, which is used to inspect the adaptability of \textit{FICGen} to mild degradations. Collectively, these datasets span a broad spectrum of degradation types, from severe low-light, underwater, aerial, and adverse weather conditions (\ie, rain, fog, snow, sandstorms) to mild blur, thereby enabling a thorough evaluation of the generalizability and robustness of \textit{FICGen} in real-world degraded scenarios.

\section{Implementation Details}
\label{sec2_}
As mentioned in the main paper, our \textit{FICGen} is built upon the pre-trained Stable Diffusion model (v1.5)~\cite{rombach2022high} and is incorporated into the mid-level ($8\times8$) feature layers and the lowest-resolution ($16\times16$) decoder layers of the denoising U-Net. All images are processed at a fixed resolution of $512\times512$ during the training and inference phases. To evaluate its generative capacity for densely distributed instances in degraded contexts, we restrict each image to at most $N=15$ objects, in line with the AeroGen setting~\cite{tang2024aerogen}. For a fair comparison, all competing L2I methods (\ie, MIGC~\cite{zhou2024migc}, CC-Diff~\cite{zhang2024cc}) are trained under identical configurations (\ie, learning rate, training epochs, and the number of instances per image).

For \textbf{fidelity evaluation}, the same number of real and synthetic images are resized to a fixed resolution of 512$\times$512, we then employ the library ``torch-fidelity''~\cite{obukhov2020torchfidelity} to compute the FID scores based on a pre-trained Inception-V3 model. It should be noted that, to ensure a fair comparison with AeroGen (CVPR 2025), the FID scores for remote sensing scenarios are computed using an Inception-V3 model fine-tuned on the RSICD~\cite{lu2017exploring} dataset, following the procedure in~\cite{xu2023txt2img}. For \textbf{alignment evaluation}, we utilize pre-trained downstream detectors, \ie, Faster R-CNN~\cite{ren2015faster} and Cascade R-CNN~\cite{cai2018cascade}, to predict detection results, which are compared against ground-truth bounding boxes and classes to compute alignment metrics. For \textbf{trainability evaluation}, we construct a synthetic training set equal in size to the real dataset, which serves as an auxiliary resource to enhance downstream detectors. To be specific, ground-truth bounding boxes (bboxes) from the training split of the degraded datasets are used as layout conditions for generating synthetic images. Following~\cite{chen2023geodiffusion}, we first discard bboxes smaller than 0.2\% of the image area, then apply data augmentation by randomly flipping bboxes with a probability of 0.8 and shifting them within 128 pixels. The generated synthetic images are combined with the real ones across various training settings. All experiments follow the default training and testing protocols of MMDetection 2.25.3~\cite{mmdetection}, with all images uniformly resized to $512\times512$ and trained under a standard $1\times$ schedule.

\begin{table}
\centering
\resizebox{0.46\textwidth}{!}{
\begin{tabular}{ccccccc}
\toprule[1.0pt]%
{Dataset} & {Train} & {Test} & {Total}&{Instances}&{Classes}&{\makecell{Degradation\\mode}}\\
\hline
ExDARK~\cite{loh2019getting} & 5,145&2,218&7,363&23,710&12&\textcolor{blue}{low-light}\\
RUOD~\cite{fu2023rethinking} & 9,800&4,200&14,000&74,904&10&\textcolor{purple}{underwater}\\
DIOR~\cite{li2020object} & 5,862&11,738&17,600&192,472&20&\textcolor{red}{aerial}\\
DAWN~\cite{kenk2020dawn} & 5,544&410&5,954&43,869&5&\textcolor{green}{adverse weather}\\
VOC 2012~\cite{everingham2010pascal} & 10,582&1,449&12,031&33,149&20&\textcolor{orange}{blur}\\
\bottomrule[1.0pt]
\end{tabular}
}
\caption{An overview of five degraded datasets.}
\label{table7}
\end{table}

\section{Additional Experimental Results}
\label{sec3_}
\subsection{More Quantitative Evaluations}
\textbf{Alignment.} Tab.~\ref{tabalign} reports a comprehensive evaluation of generative fidelity (FID) and alignment (detection AP) across \textbf{five} degraded benchmarks (DIOR-H~\cite{li2020object}, ExDARK~\cite{loh2019getting}, RUOD~\cite{fu2023rethinking}, DAWN~\cite{kenk2020dawn}, and blurred VOC 2012~\cite{everingham2010pascal}) and one natural benchmark (VOC 2012~\cite{everingham2010pascal}). The alignment is assessed using two pre-trained detectors, Faster R-CNN (R50)~\cite{ren2015faster} and Cascade R-CNN (R50)~\cite{cai2018cascade}, enabling a fair comparison with prior L2I approaches. Overall, our FICGen consistently delivers superior alignment and fidelity, attaining a \textbf{27.6 mAP} on DIOR-H, outperforming MIGC (21.8) and CC-Diff (23.6) by \textbf{5.8} and \textbf{4.0} points, respectively. The gains are particularly pronounced on challenging semantic categories, such as ``\textit{windmill}'' (16.0 \textit{vs.} 3.4/5.6) and ``\textit{airplane}'' (23.8 \textit{vs.} 6.8/9.5). 

\begin{table*}
\centering
\resizebox{1.0\textwidth}{!}{
\begin{tabular}{ccc|cccccc|ccc}
\toprule[1.2pt]%
Detector& Method &FID$\downarrow$& \multicolumn{6}{|c|}{Object Detection (AP) for Sampled Classes /\%} & mAP$\uparrow$ & AP\_{50}$\uparrow$ &AP\_{75}$\uparrow$ \\

\hline \rowcolor{gray!5}
\multicolumn{3}{c}{\textcolor{red}{\textbf{DIOR-H}~\cite{li2020object}}}&windmill&airport&stadium&ballfield&airplane&golffield& & & \\
Faster R-CNN & Oracle&-&29.0&32.2&26.7&50.2&33.5&34.7&33.4&55.6&35.0\\
\hline
Faster R-CNN & MIGC~\cite{zhou2024migc}&31.64&3.4&\textcolor{red}{33.6}&\underline{18.0}&32.4&6.8&43.8&21.8&38.4&17.5\\
Faster R-CNN & CC-Diff~\cite{zhang2024cc}&\textcolor{red}{30.88}&5.6&31.4&\textcolor{red}{21.4}&\underline{44.0}&\underline{9.5}&\underline{50.6}&\underline{23.6}&\underline{42.4}&\underline{21.4}\\
\rowcolor{gray!10}
Faster R-CNN& FICGen (ours)&\underline{31.25}&\textcolor{red}{16.0}&\underline{33.5}&16.9&\textcolor{red}{50.0}&\textcolor{red}{23.8}&\textcolor{red}{51.2}&\textcolor{red}{27.6}&\textcolor{red}{48.7}&\textcolor{red}{27.6}\\

\hline \rowcolor{gray!5}
\multicolumn{3}{c}{\textcolor{purple}{\textbf{RUOD}~\cite{fu2023rethinking}}}&starfish&echinus&fish&scallop&corals&turtle& & &\\
Faster R-CNN& Oracle&-&45.9&44.1&41.9&37.5&45.9&67.8&50.5&80.2&54.4\\
\hline
Faster R-CNN & MIGC~\cite{zhou2024migc}&26.50&20.6&12.2&14.9&\underline{13.5}&22.8&48.2&27.2&54.1&24.6\\
Faster R-CNN&CC-Diff~\cite{zhang2024cc}&\underline{25.21}&\underline{22.7}&\underline{13.3}&\underline{16.8}&12.5&25.6&50.7&\underline{29.7}&\underline{58.4}&\underline{27.9}\\
\rowcolor{gray!10}
Faster R-CNN&FICGen (ours)&\textcolor{red}{25.10}&\textcolor{red}{32.1}&\textcolor{red}{28.5}&\textcolor{red}{25.0}&\textcolor{red}{23.7}&\textcolor{red}{33.7}&\textcolor{red}{52.5}&\textcolor{red}{37.0}&\textcolor{red}{68.6}&\textcolor{red}{36.5}\\

\hline \rowcolor{gray!5}
\multicolumn{3}{c}{\textcolor{orange}{\textbf{blurred VOC 2012}~\cite{everingham2010pascal}}}&aeroplane&boat&cow&bottle&sheep&person& & &\\
Faster R-CNN& Oracle&-&43.7&26.9&24.3&19.7&27.9&34.5&31.5&56.5&32.4\\
\hline
Faster R-CNN & MIGC~\cite{zhou2024migc}&62.66&42.4&27.0&23.9&\underline{20.2}&21.8&30.3&34.7&65.8&33.1\\
Faster R-CNN&CC-Diff~\cite{zhang2024cc}&\underline{62.20}&\underline{44.0}&\underline{28.6}&\underline{26.5}&19.6&\underline{27.1}&\underline{32.5}&\underline{36.7}&\underline{67.6}&\underline{36.3}\\
\rowcolor{gray!10}
Faster R-CNN&FICGen (ours)&\textcolor{red}{58.02}&\textcolor{red}{49.0}&\textcolor{red}{36.1}&\textcolor{red}{36.2}&\textcolor{red}{23.2}&\textcolor{red}{35.4}&\textcolor{red}{35.4}&\textcolor{red}{40.7}&\textcolor{red}{70.3}&\textcolor{red}{42.7}\\

\hline \rowcolor{gray!5}
\multicolumn{3}{c}{\textcolor{cyan}{\textbf{VOC 2012}~\cite{everingham2010pascal}}}&aeroplane&boat&cow&bottle&sheep&person& & &\\
Faster R-CNN& Oracle&-&57.1&42.9&48.7&33.2&46.1&48.1&48.3&76.8&52.5\\
\hline
Faster R-CNN & AeroGen~\cite{tang2024aerogen}&\textcolor{red}{45.21}&38.1&27.7&30.9&28.7&35.4&29.5&36.8&65.7&36.4\\
Faster R-CNN & MIGC~\cite{zhou2024migc}&50.60&47.3&33.9&44.9&24.1&35.0&34.3&45.1&78.5&47.4\\
Faster R-CNN&CC-Diff~\cite{zhang2024cc}&\underline{48.70}&\underline{55.0}&\underline{36.2}&\underline{46.4}&\underline{27.2}&\underline{37.8}&\underline{35.5}&\underline{47.9}&\underline{79.1}&\underline{52.0}\\
\rowcolor{gray!10}
Faster R-CNN&FICGen (ours)&48.93&\textcolor{red}{57.7}&\textcolor{red}{47.9}&\textcolor{red}{49.9}&\textcolor{red}{37.4}&\textcolor{red}{45.9}&\textcolor{red}{44.6}&\textcolor{red}{54.2}&\textcolor{red}{83.5}&\textcolor{red}{60.2}\\

\hline \rowcolor{gray!5}
\multicolumn{3}{c}{\textcolor{blue}{\textbf{ExDARK}~\cite{loh2019getting}}}&bicycle&motorbike&boat&car&cup&bottle& & &\\
Cascade R-CNN & Oracle&-&48.5&33.6&30.4&40.2&29.0&32.1&37.2&65.8&37.8\\
\hline
Cascade R-CNN& MIGC~\cite{zhou2024migc}&45.76&45.9&25.4&26.6&28.8&21.9&21.5&32.4&63.5&29.5\\
Cascade R-CNN& CC-Diff~\cite{zhang2024cc}&\underline{44.26}&\underline{48.5}&\underline{32.5}&\textcolor{red}{30.6}&\underline{33.0}&\underline{23.3}&\underline{25.9}&\underline{35.1}&\underline{65.6}&\underline{34.1}\\
\rowcolor{gray!10}
Cascade R-CNN& FICGen (ours)&\textcolor{red}{42.40}&\textcolor{red}{57.3}&\textcolor{red}{38.5}&\underline{29.4}&\textcolor{red}{43.9}&\textcolor{red}{35.8}&\textcolor{red}{34.8}&\textcolor{red}{42.5}&\textcolor{red}{73.0}&\textcolor{red}{45.1}\\
\hline \rowcolor{gray!5}
\multicolumn{3}{c}{\textcolor{green}{\textbf{DAWN}~\cite{kenk2020dawn}}}&motorcycle&person&bus&truck&car&-& & &\\
Cascade R-CNN& Oracle&-&19.4&20.6&27.0&26.7&42.8&-&27.3&46.4&26.3\\
\hline
Cascade R-CNN& MIGC~\cite{zhou2024migc}&70.10&9.3&7.2&17.0&14.6&14.8&-&12.6&32.3&8.6 \\             
Cascade R-CNN& CC-Diff~\cite{zhang2024cc}&\underline{68.56}&\underline{13.2}&\underline{9.4}&\underline{19.4}&\underline{17.6}&\underline{19.0}&-&\underline{15.7}&\underline{33.9}&\underline{14.8}  
\\
\rowcolor{gray!10}
Cascade R-CNN& FICGen (ours)&\textcolor{red}{68.31}&\textcolor{red}{25.3}&\textcolor{red}{20.3}&\textcolor{red}{19.8}&\textcolor{red}{19.6}&\textcolor{red}{28.2}&-&\textcolor{red}{22.6}&\textcolor{red}{44.3}&\textcolor{red}{21.5}\\
\bottomrule[1.2pt]
\end{tabular}
}
\caption{Quantitative comparison of generative fidelity and alignment on five degraded datasets (DIOR-H~\cite{li2020object}, ExDARK~\cite{loh2019getting}, RUOD~\cite{fu2023rethinking}, DAWN~\cite{kenk2020dawn}, and blurred VOC 2012~\cite{everingham2010pascal}) and one natural image dataset (VOC 2012~\cite{everingham2010pascal}). The performance is evaluated using off-the-shelf detectors Faster R-CNN (R50)~\cite{ren2015faster} and Cascade R-CNN (R50)~\cite{cai2018cascade} on synthetic test images generated by different L2I methods. ``\textit{Oracle}'' denotes the real test set baseline (\ie, upper bound). The top-2 performers are marked in \textcolor{red}{red} and \underline{underlined}.}
\label{tabalign}
\end{table*}

Similar improvements are observed across other benchmarks. On RUOD, which involves dense underwater instances, FICGen achieves a \textbf{37.0 mAP}, surpassing CC-Diff by \textbf{7.3} and MIGC by \textbf{9.8}, owing to its frequency-inspired contextual disentanglement that mitigates the submersion of underwater objects against homogeneous aquatic backgrounds. On blurred VOC 2012, our method elevates AP for almost all categories such as ``\textit{cow}'' (36.2 \textit{vs.} 23.9/26.5) and ``\textit{sheep}'' (35.4 \textit{vs.} 21.8/27.1), resulting in a \textbf{40.7 mAP}, which is a clear lead over CC-Diff (36.7) and MIGC (34.7). For ExDARK and DAWN, which represent extreme low-light and adverse-weather conditions, FICGen delivers notable boosts in categories with attenuated high-frequency cues, such as ``\textit{motorbike}'' (38.5 \textit{vs.} 25.4/32.5) and ``\textit{motorcycle}'' (25.3 \textit{vs.} 9.3/13.2), achieving \textbf{42.5} and \textbf{22.6 mAP}, respectively.

Crucially, these improvements can be attributed to FICGen’s contextual disentanglement mechanism, which effectively separates high-frequency instance details from the dominant low-frequency surroundings, while simultaneously preserving essential degraded contextual characteristics such as illumination and texture.
\begin{figure*}
\centering
\includegraphics[width=1.00\textwidth]{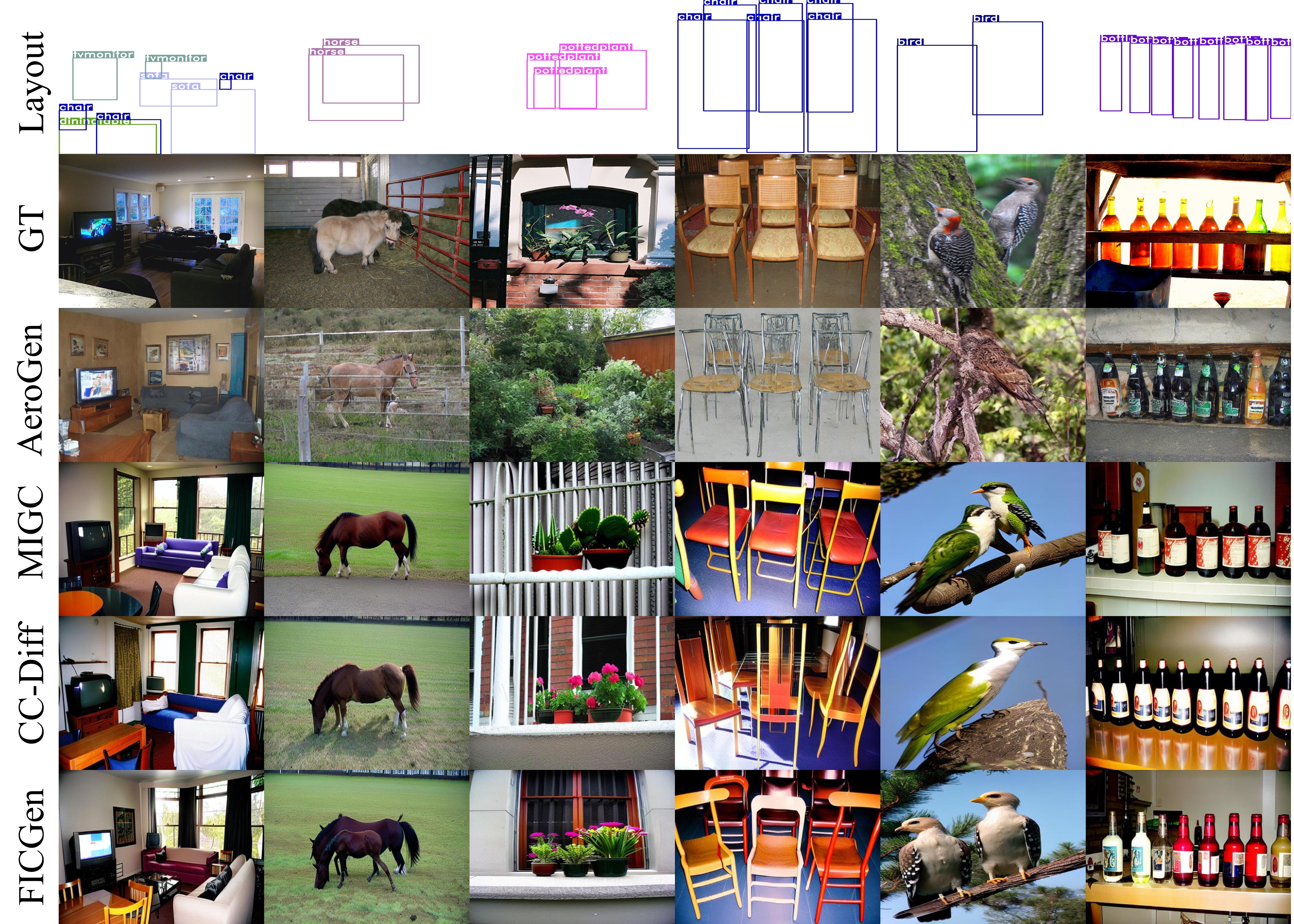}
\caption{Qualitative comparison of natural images (VOC 2012~\cite{everingham2010pascal}) generated by different L2I methods. Zoom in for more detail.}
\label{FIG:fuluvoccom}
\end{figure*}

\textbf{Trainability.} Following the trainability protocol in~\cite{chen2023geodiffusion}, we leverage ground-truth bounding boxes from the degraded datasets as layout inputs to synthesize additional training images, which are combined with real samples to effectively double the training set size. As summarized in Tabs.~\ref{tabexdark}–~\ref{tabvoc}, our proposed FICGen consistently delivers superior gains in downstream detection performance across both degraded and natural scenarios. On the ExDARK benchmark, FICGen achieves the highest mean AP, with notable improvements for semantic categories such as ``\textit{motorbike}'' (\textbf{+3.8}) relative to the Oracle upper bound and ``\textit{people}'' (\textbf{+1.6}) over the closest CC-Diff. On DAWN, despite the small scale of available data, FICGen yields competitive boosts, including \textbf{+4.9} for ``\textit{motorcycle}'' and \textbf{+4.4} for ``\textit{bus}.'' More importantly, on natural VOC 2012, which reflects the generalization capability, FICGen achieves the best overall mean AP (\textbf{50.5}), surpassing CC-Diff by \textbf{+0.9} and MIGC by \textbf{+1.1}. These downstream trainability results confirm that the degraded images synthesized by FICGen not only alleviate the scarcity of training data in adverse conditions but also serve as effective auxiliary resources to enhance the accuracy of downstream detectors. 
\begin{table*}
\centering
\resizebox{1.0\textwidth}{!}{
\begin{tabular}{ccccc|ccccc}
\toprule[1.5pt]%
category & Oracle & MIGC~\cite{zhou2024migc}&CC-Diff~\cite{zhang2024cc} & FICGen & category & Oracle & MIGC~\cite{zhou2024migc}&CC-Diff~\cite{zhang2024cc} & FICGen \\
\hline
bicycle & 46.4 & 48.3 &49.6& 47.8 & chair & 25.4 & 26.8&26.8 & 27.5 \\
boat & 30.2 & 31.1 & 29.7&31.7 & cup & 28.6 & 28.2 &28.7& 29.6 \\
bottle & 31.8 & 31.6 &31.1& 32.8 & dog & 44.1 & 47.9 &47.6& 48.5 \\
bus & 56.8 & 60.8 & 59.8&59.9 & motorbike & 33.3 & 34.4 &34.1& 37.1 \\
car & 38.1 & 38.3 & 38.4&38.8 & people & 32.9 & 32.5 &32.4& 34.0 \\
cat & 41.8 & 41.4 & 42.0&42.4 & table & 20.6 & 24.1 & 24.1&22.5 \\
\hline
&&&&&All (\textbf{mAP \%}) &35.8&37.1&37.0&37.7\\
\bottomrule[1.5pt]
\end{tabular}
}
\caption{Detection accuracy (\%) per class on the ExDARK \textit{test} set using Faster R-CNN (R50)~\cite{ren2015faster}, trained on 5.1k real and 5.1k synthetic images generated by different L2I methods.}\label{tabexdark}
\end{table*}

\begin{table}
\centering
\resizebox{0.5\textwidth}{!}{
\begin{tabular}{ccccc}
\toprule[1.5pt]%
category & Oracle & MIGC~\cite{zhou2024migc}&CC-Diff~\cite{zhang2024cc} & FICGen \\
\hline
motorcycle & 18.0 & 23.0 &20.0& 22.9\\
person & 19.6 & 19.5 & 19.5&20.2\\
bus & 21.8 & 21.6 &27.8& 26.2\\
truck & 23.4 & 19.6 & 21.6&20.5\\
car & 41.1 & 39.4 & 40.0&39.6\\
\hline
All (\textbf{mAP \%}) &24.8&24.6&25.8&25.9\\
\bottomrule[1.5pt]
\end{tabular}
}
\caption{Detection accuracy (\%) per class on the DAWN \textit{test} set using Faster R-CNN (R50)~\cite{ren2015faster}, trained on 5.5k real and 5.5k synthetic images generated by different L2I methods.}
\label{tabledawn}
\end{table}

\begin{table*}
\centering
\resizebox{1.0\textwidth}{!}{
\begin{tabular}{ccccc|ccccc}
\toprule[1.5pt]%
category & Oracle & MIGC~\cite{zhou2024migc}&CC-Diff~\cite{zhang2024cc} & FICGen & category & Oracle & MIGC~\cite{zhou2024migc}&CC-Diff~\cite{zhang2024cc} & FICGen \\
\hline
aeroplane & 57.1 & 58.0 &59.6& 59.2 & diningtable & 42.8 & 44.5&42.3 & 45.2 \\
bicycle & 51.6 & 54.3 & 53.1&53.6 & dog & 59.6 & 59.8 &60.9& 61.9 \\
bird & 52.2 & 53.6 &54.1& 55.7 & horse & 54.9 & 55.7 &55.2& 55.4 \\
boat & 42.9 & 42.5 & 44.4&47.3 & motorbike & 56.2 & 58.7 &59.1& 57.8 \\
bottle & 33.2 & 34.7 & 36.3&34.9 & person & 48.1 & 47.9 &48.1& 48.9 \\
bus & 59.4 & 60.1 & 60.1&62.2 & pottedplant & 26.9 & 25.8 & 26.0&27.9 \\
car & 43.0 & 43.6 & 44.6&45.9 & sheep & 46.1 & 46.7 & 45.7&47.3 \\
cat & 62.2 & 61.5 & 62.1&63.0 & sofa & 43.6 & 47.9 & 47.5&47.6 \\
chair & 26.5 & 28.7 & 28.9&29.6 & train & 59.9 & 61.4 & 62.0&61.6 \\
cow & 48.7 & 51.6 & 51.0&51.0 & tvmonitor & 51.4 & 50.9 & 50.7&53.1 \\
\hline
&&&&&All (\textbf{mAP \%}) &48.3&49.4&49.6&50.5\\
\bottomrule[1.5pt]
\end{tabular}
}
\caption{Detection accuracy (\%) per class on natural VOC 2012 \textit{test} set using Faster R-CNN (R50)~\cite{ren2015faster}, trained on 10.6k real and 10.6k synthetic images generated by different L2I methods.}\label{tabvoc}
\end{table*}

\begin{table*}
\centering
\resizebox{1\textwidth}{!}{
\renewcommand{\arraystretch}{0.9}
\begin{tabular}{cc|ccc|ccc|ccc}
\toprule[1.0pt]%
\multirow{2}*{Dataset} & \multirow{2}*{Method}&  \multicolumn{3}{c|}{Sparse  \small(\textcolor{red}{33\%}/\textcolor{blue}{39\%}/\textcolor{purple}{23\%}/\textcolor{green}{27\%})}&\multicolumn{3}{c|}{Partial \small(\textcolor{red}{21\%}/\textcolor{blue}{15\%}/\textcolor{purple}{18\%}/\textcolor{green}{18\%})}&\multicolumn{3}{c}{Heavy \small(\textcolor{red}{46\%}/\textcolor{blue}{46\%}/\textcolor{purple}{59\%}/\textcolor{green}{55\%})}\\
\cmidrule{3-11}%
&&mAP&AP\_{50}&AP\_{75}&mAP&AP\_{50}&AP\_{75}&mAP&AP\_{50}&AP\_{75}\\
\hline
\multirow{4}{*}{\textcolor{red}{DIOR-H}~\cite{li2020object}}
&Oracle&45.9&70.2&50.9&35.2&53.3&38.6&28.7&47.6&30.7\\
\hline
&MIGC~\cite{zhou2024migc}&33.1&64.4&30.7&19.1&38.8&16.7&16.1&33.9&12.9\\
&CC-Diff~\cite{zhang2024cc}&\underline{38.7}&\underline{66.3}&\underline{40.7}&\underline{24.4}&\underline{42.9}&\underline{25.0}&\underline{19.5}&\underline{36.5}&\underline{19.1}\\
\rowcolor{gray!10}
&FICGen&\textcolor{red}{44.8}&\textcolor{red}{72.4}&\textcolor{red}{48.8}&\textcolor{red}{30.1}&\textcolor{red}{49.2}&\textcolor{red}{32.5}&\textcolor{red}{24.1}&\textcolor{red}{42.6}&\textcolor{red}{24.9}\\
\hline
\multirow{4}{*}{\textcolor{blue}{ExDARK}~\cite{loh2019getting}}
&Oracle&55.6&87.4&65.7&42.6&73.4&46.3&28.7&58.6&25.3\\
\hline
&MIGC~\cite{zhou2024migc}&59.3&\underline{95.1}&68.0&44.7&85.8&43.1&24.9&58.5&16.7\\
&CC-Diff~\cite{zhang2024cc}&\underline{60.6}&95.1&\underline{72.2}&\underline{47.7}&\underline{84.3}&\underline{49.0}&\underline{26.1}&\underline{58.5}&\underline{19.3}\\
\rowcolor{gray!10}
&FICGen&\textcolor{red}{63.8}&\textcolor{red}{95.3}&\textcolor{red}{75.8}&\textcolor{red}{54.0}&\textcolor{red}{89.0}&\textcolor{red}{59.8}&\textcolor{red}{32.1}&\textcolor{red}{65.3}&\textcolor{red}{28.0}\\
\hline
\multirow{4}{*}{\textcolor{purple}{RUOD}~\cite{fu2023rethinking}}
&Oracle&56.2&81.2&64.3&53.0&81.6&58.5&48.1&79.6&51.0\\
\hline
&MIGC~\cite{zhou2024migc}&40.1&75.2&39.1&30.4&63.7&25.8&17.8&39.5&13.4\\
&CC-Diff~\cite{zhang2024cc}&\underline{47.8}&\underline{80.6}&\underline{51.3}&\underline{38.2}&\underline{73.1}&\underline{38.3}&\underline{26.3}&\underline{54.2}&\underline{23.0}\\
\rowcolor{gray!10}
&FICGen&\textcolor{red}{54.3}&\textcolor{red}{85.6}&\textcolor{red}{62.2}&\textcolor{red}{46.5}&\textcolor{red}{80.4}&\textcolor{red}{49.6}&\textcolor{red}{33.5}&\textcolor{red}{65.3}&\textcolor{red}{31.1}\\
\hline
\multirow{4}{*}{\textcolor{green}{DAWN}~\cite{kenk2020dawn}}
&Oracle&36.1&56.5&35.2&34.9&56.3&42.9&23.8&45.3&21.4\\
\hline
&MIGC~\cite{zhou2024migc}&25.3&52.3&21.3&17.8&38.8&13.0&11.1&29.9&5.7\\
&CC-Diff~\cite{zhang2024cc}&\underline{38.2}&\underline{76.7}&\underline{33.6}&\underline{26.4}&\underline{50.9}&\underline{26.3}&\underline{13.2}&\underline{32.1}&\underline{9.2}\\
\rowcolor{gray!10}
&FICGen&\textcolor{red}{44.5}&\textcolor{red}{71.7}&\textcolor{red}{48.1}&\textcolor{red}{34.5}&\textcolor{red}{54.0}&\textcolor{red}{40.7}&\textcolor{red}{20.0}&\textcolor{red}{40.8}&\textcolor{red}{16.1}\\
\bottomrule[1.0pt]
\end{tabular}
}
\caption{Quantitative comparison of alignment across four datasets under three occlusion degrees, with detection evaluated by Faster R-CNN (R50). The numbers in brackets represent the proportion of each occlusion level within the dataset.}
\label{tabocc}
\end{table*}

\begin{table}
\centering
\resizebox{0.5\textwidth}{!}{
\scriptsize
\begin{tabular}{c|ccc|c}
\toprule[1.0pt]%
 Method & Trainable Params (M) &Inf.Time (s/img)& Inf. Mem (GB) &mAP \\
\hline
CC-Diff &$\sim$261&8&$\sim$12.2&26.4\\
AeroGen&$\sim$905&8&$\sim$13.9&29.8\\
\rowcolor{gray!10}
FICGen (ours)&$\sim$304&10&$\sim$12.6&31.2\\
\bottomrule[1.0pt]
\end{tabular}
}
\caption{Comparison of trainable parameters and inference efficiency with AeroGen~\cite{tang2024aerogen} and CC-Diff~\cite{zhang2024cc}.}
\label{tableinf}
\end{table}

\begin{figure}[!t]
\centering
\includegraphics[width=0.46\textwidth]{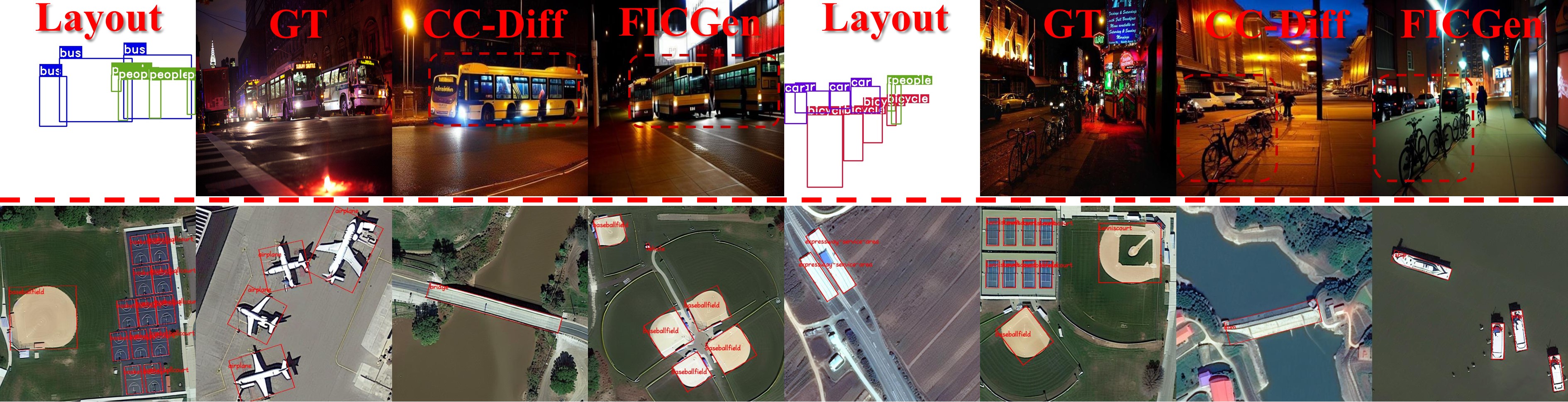}
\caption{Further qualitative comparison with CC-Diff on ExDARK and generalization to oriented layout inputs~\cite{cheng2022anchor}.}
\label{figgeneralization}
\end{figure}

\begin{figure}[!t]
\centering
\includegraphics[width=0.46\textwidth]{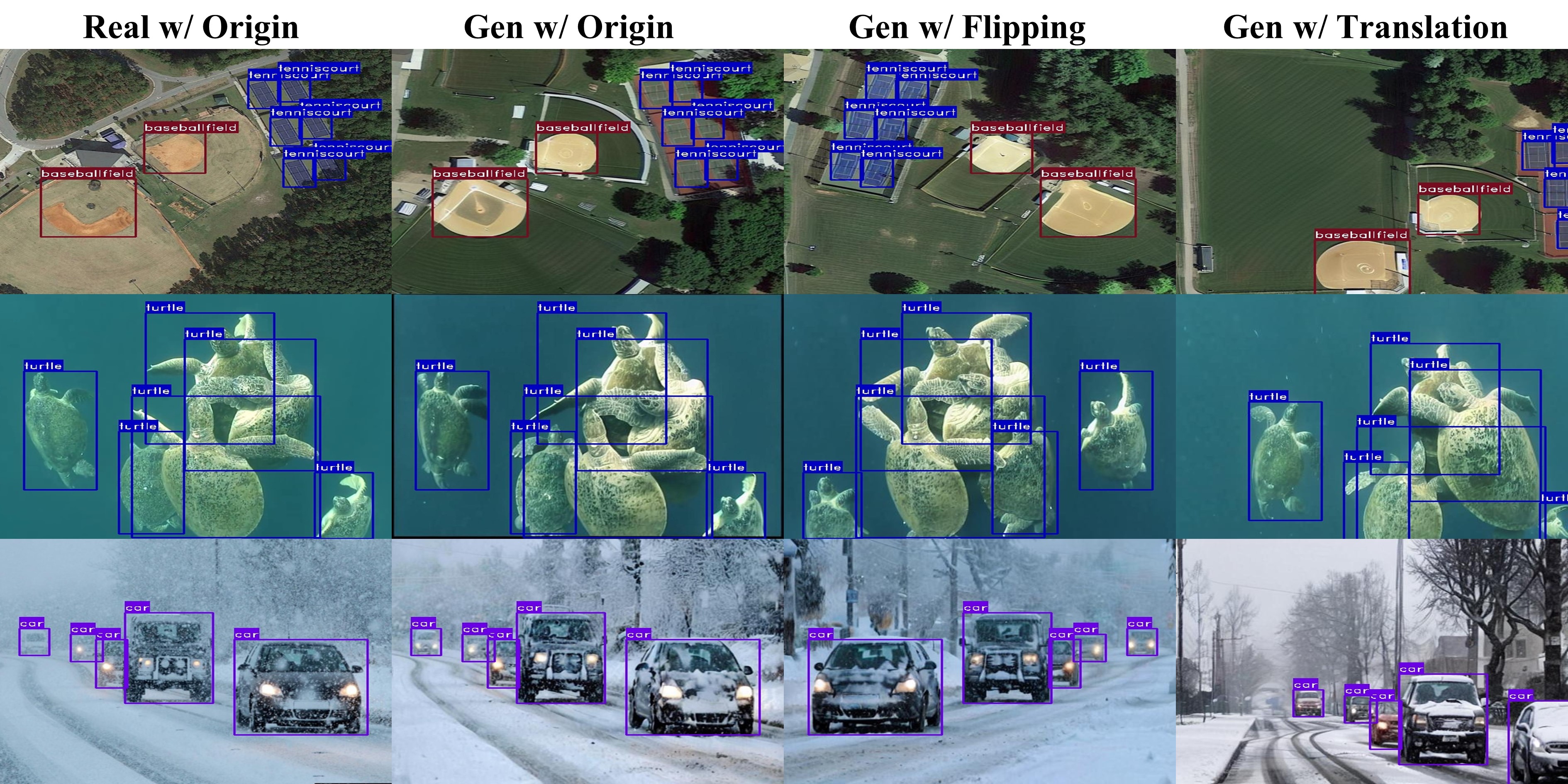}
\caption{Visualization results generated by FICGen conditioned on augmented geometric layouts.}
\label{figaugmented}
\end{figure}

\textbf{Robust Control.} To further investigate the generative robustness of FICGen towards spatially entangled instances under degraded conditions, Tab.~\ref{tabocc} presents a detailed alignment comparison on four benchmarks with varying occlusion levels. Each dataset is divided into three occlusion levels, \ie, Sparse, Partial, and Heavy, according to the number of instances and their mutual occlusion measured by Intersection over Union. A pre-trained Faster R-CNN (R50) is then used to assess the alignment between the generated instances and their corresponding layouts at each occlusion level.

As reported in Tab.~\ref{tabocc}, our FICGen demonstrates remarkable robustness across varying occlusion levels, significantly outperforming prior L2I approaches. On sparsely occluded cases, where inter-instance interference is minimal, FICGen already yields clear gains, achieving a mAP of \textbf{44.8} on DIOR-H and \textbf{63.8} on ExDARK, surpassing CC-Diff by \textbf{6.1} and \textbf{3.2}, respectively. These improvements become more pronounced under partial occlusion, where overlapping instances introduce complex spatial entanglement. For example, on RUOD, FICGen attains a \textbf{46.5 mAP}, outperforming MIGC and CC-Diff by \textbf{16.1} and \textbf{8.3}, respectively. The benefits are most substantial under heavy occlusion, where conventional methods often suffer from ``object omission and merging'' due to severe spatial collisions. On DAWN and RUOD, which contain up to \textbf{55}\% and \textbf{59}\% heavily occluded instances, FICGen secures \textbf{20.0} and \textbf{33.5 mAP}, nearly doubling the performance of MIGC (\textbf{11.1} and \textbf{17.8}) and substantially outperforming CC-Diff (\textbf{13.2} and \textbf{26.3}).  Collectively, these results underline FICGen’s ability to faithfully adhere to user-specified layouts and mitigate occlusion-induced distortions, thereby generating structurally consistent degraded images that benefit downstream perception tasks.
\begin{figure}[!t]
\centering
\includegraphics[width=0.46\textwidth]{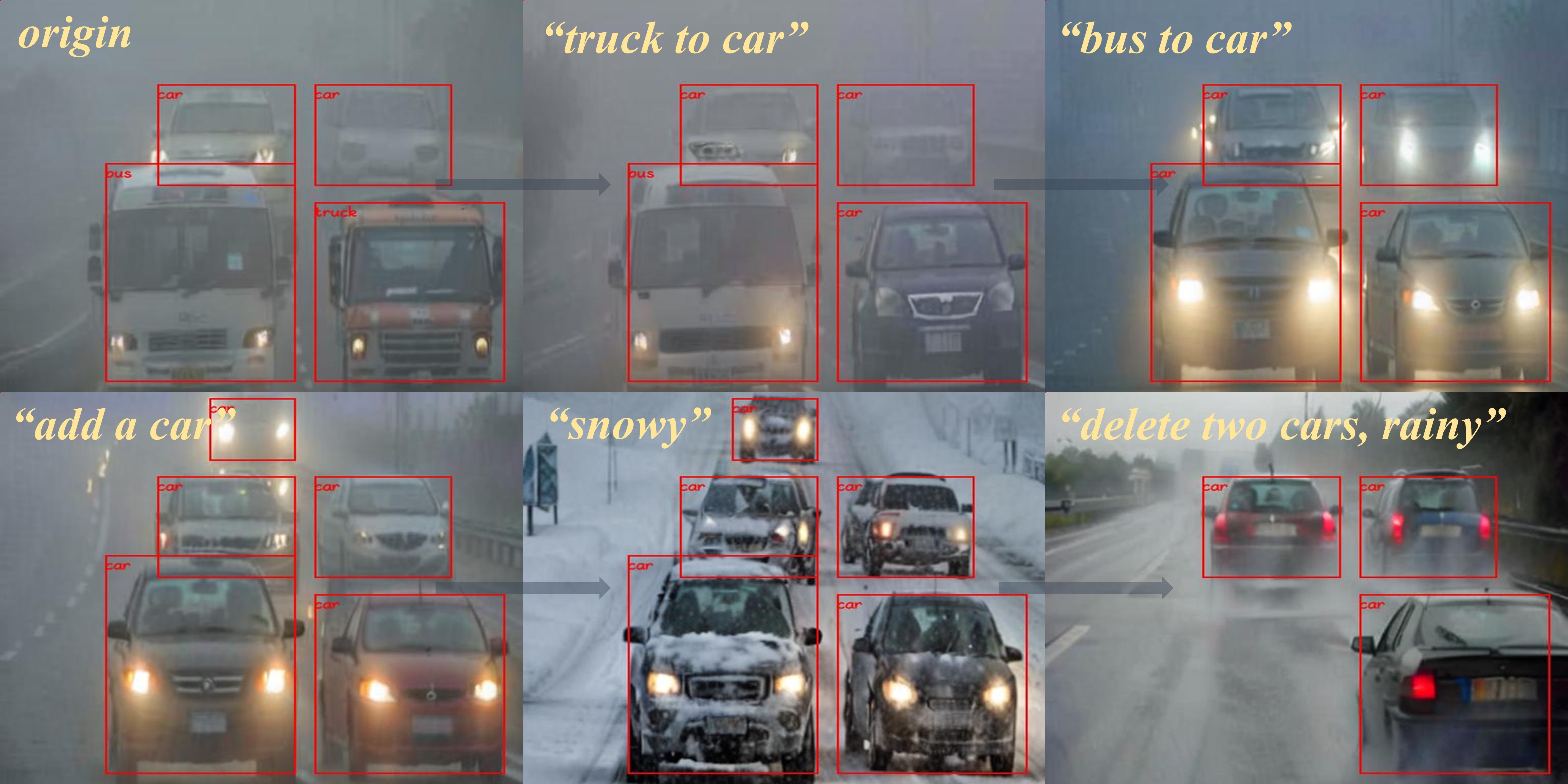}
\caption{Visualization results of the continuous generation of instance interactivity (\ie, addition, removal, transform and weather change) by our FICGen on the DAWN dataset.}
\label{figdawnedit}
\end{figure}

\textbf{Inference Efficiency Analysis.} Tab.~\ref{tableinf} reports the number of trainable parameters and the per-image inference time for identical settings on the DIOR dataset. FICGen contains only around a third of the parameters of AeroGen ($\sim$304M \textit{vs.} $\sim$905M). Compared with CC-Diff, FICGen achieves the highest mAP while maintaining a comparable overhead, requiring only an additional 40M trainable parameters and incurring a modest 2-second inference delay.
\subsection{More Qualitative Evaluations}
Fig.~\ref{FIG:fuluvoccom} presents a qualitative comparison of generation results produced by different L2I methods on natural scenes from VOC 2012. Existing approaches exhibit notable limitations in generative quality, often suffering from hallucination artifacts such as missing objects and incorrect merging. For instance, AeroGen generates synthesized images whose visual characteristics deviate substantially from real-world counterparts, while MIGC and CC-Diff frequently merge adjacent objects, such as \textbf{two horses being fused into one (second column) or three potted plants collapsing into two or overflowing excessively (third column)}. Moreover, when processing densely distributed objects, such as the array of bottles in column six, these methods often fail to preserve accurate spatial arrangement and object counts. In contrast, although primarily designed for degraded image generation, our FICGen demonstrates superior performance in natural scenarios, particularly in maintaining object quantity, spatial positioning, and scale. Specifically, it accurately reproduces \textbf{highly overlapping objects, including the two horses in the second column and six closely packed chairs in the fourth column, while preserving fidelity for small-scale targets, exemplified by the chair in the first column.}

In the additional low-light results shown in Fig.~\ref{figgeneralization}, CC-Diff \textbf{suffers from the illusion of merging three ``\textit{buses}'' into one and omitting dense ``\textit{motorbike}'' instances}, whereas our FICGen effectively resolves such contextual illusion issues. What's more, as shown in the second row of Fig.~\ref{figgeneralization}, FICGen exhibits strong adaptability to oriented layout controls, ensuring consistent generation quality under such constraints.

Figs.~\ref{figaugmented} and ~\ref{figdawnedit} present the generative results on layout manipulation. Fig.~\ref{figaugmented} demonstrates that FICGen maintains robust generation for augmented layouts, such as flipping and translation, while Fig.~\ref{figdawnedit} shows that FICGen consistently produces coherent content under continuous layout modifications, including object addition (\ie, ``\textit{add a car}''), removal (\ie, ``\textit{delete two cars}''), categorical transformation (\ie, ``\textit{truck to car}''), and weather changes (\ie, ``\textit{snowy, rainy}''), highlighting its flexibility and user-controllable interactivity. These qualitative results further underscore the advantages of FICGen in degraded image generation. 

Fig.~\ref{figdetection} presents the detection results of Faster R-CNN \textit{wi/wo} FICGen for auxiliary training. The synthesized degraded images from FICGen notably improve the detector’s localization accuracy, particularly for remote sensing instances such as ``\textit{baseballfield}'' and ``\textit{people}'' in low-light conditions.

\subsection{More Architecture Details}

\begin{figure}[!t]
\centering
\includegraphics[width=0.48\textwidth]{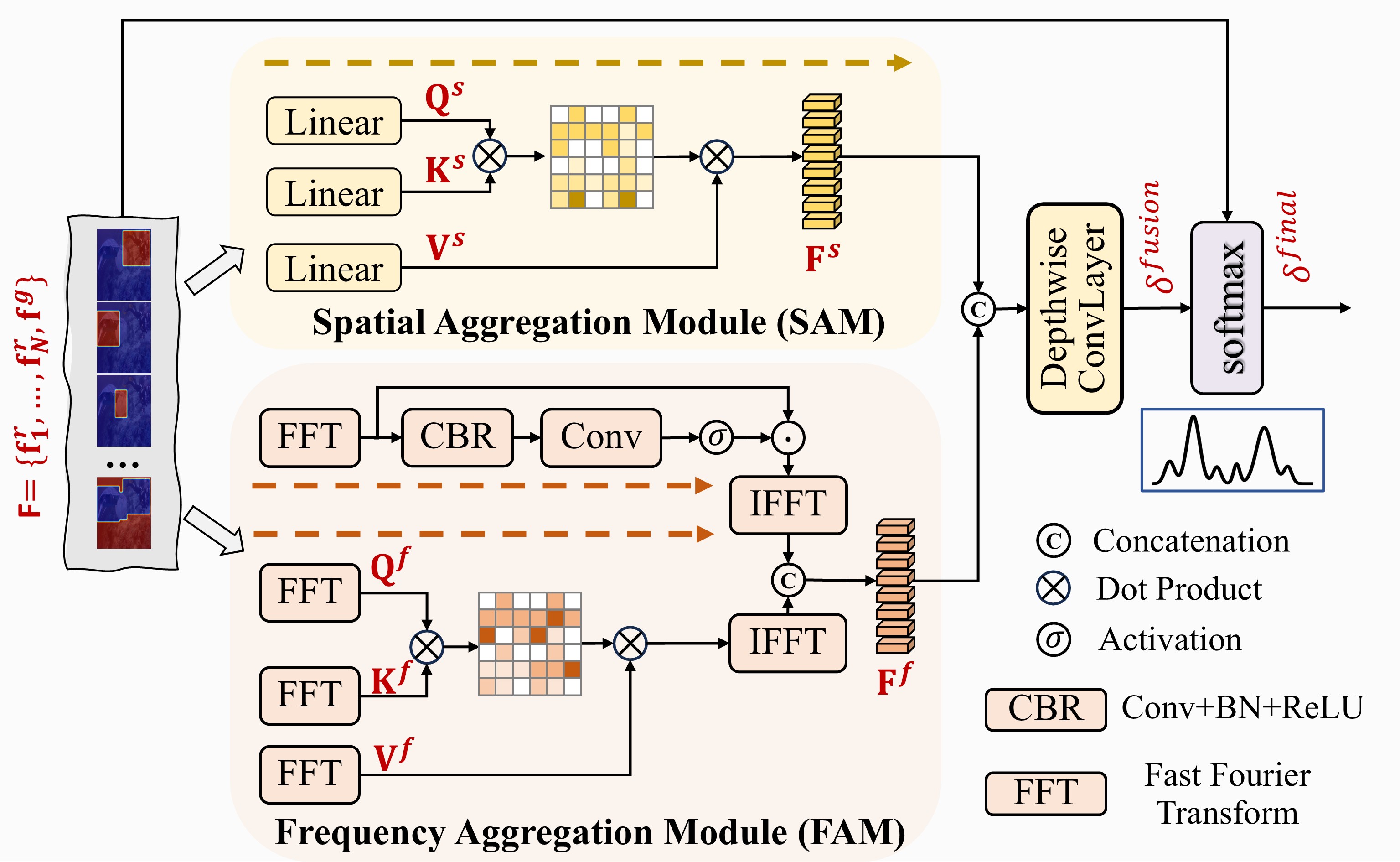}
\caption{Architectural details of the proposed Adaptive Spatial-Frequency Aggregation Module (ASFA).}
\label{figasfa}
\end{figure}

\begin{table}
\centering
\resizebox{0.48\textwidth}{!}{
\begin{tabular}{ccccc}
\toprule[1.0pt]%
Fusion & airplane&mAP&AP\_{50}&AP\_{75}\\
\hline
Depth-wise&23.8&27.6&48.7&27.6\\
Point-wise&16.0&26.4&48.1&26.4\\
\bottomrule[1.0pt]
\end{tabular}
}
\caption{The ablation results of different fusion strategies for SAM and FAM outputs.}
\label{tableasfa}
\end{table}
Fig.~\ref{figasfa} illustrates the architectural details of the proposed Adaptive Spatial-Frequency Aggregation (ASFA) module. Inspired by~\cite{sun2024frequency}, we adopt a dual-branch spatial-frequency aggregation strategy to integrate the disentangled degraded representations in the latent space. The spatial branch captures contextual dependencies, including semantic correlations and degradation similarities among various objects. Concurrently, the frequency branch focuses on fine-grained attributes such as edge structures and texture details. Next, we further fuse the dual-stream outputs of the SAM and FAM at a lower cost by using a single-layer depthwise separable convolution to enhance local perception within different degraded regions. Finally, adaptive weights for context-aware aggregation are obtained via a softmax operation. As shown in Tab.~\ref{tableasfa}, this fusion strategy substantially outperforms the point-wise alternative, particularly for small objects like “\textit{airplane}”, where AP improves from 16.0 to 23.8. These results highlight the effectiveness of our adaptive aggregation in capturing both global dependencies and local structural cues under complex degradation scenarios.

\begin{figure}[!t]
\centering
\includegraphics[width=0.46\textwidth]{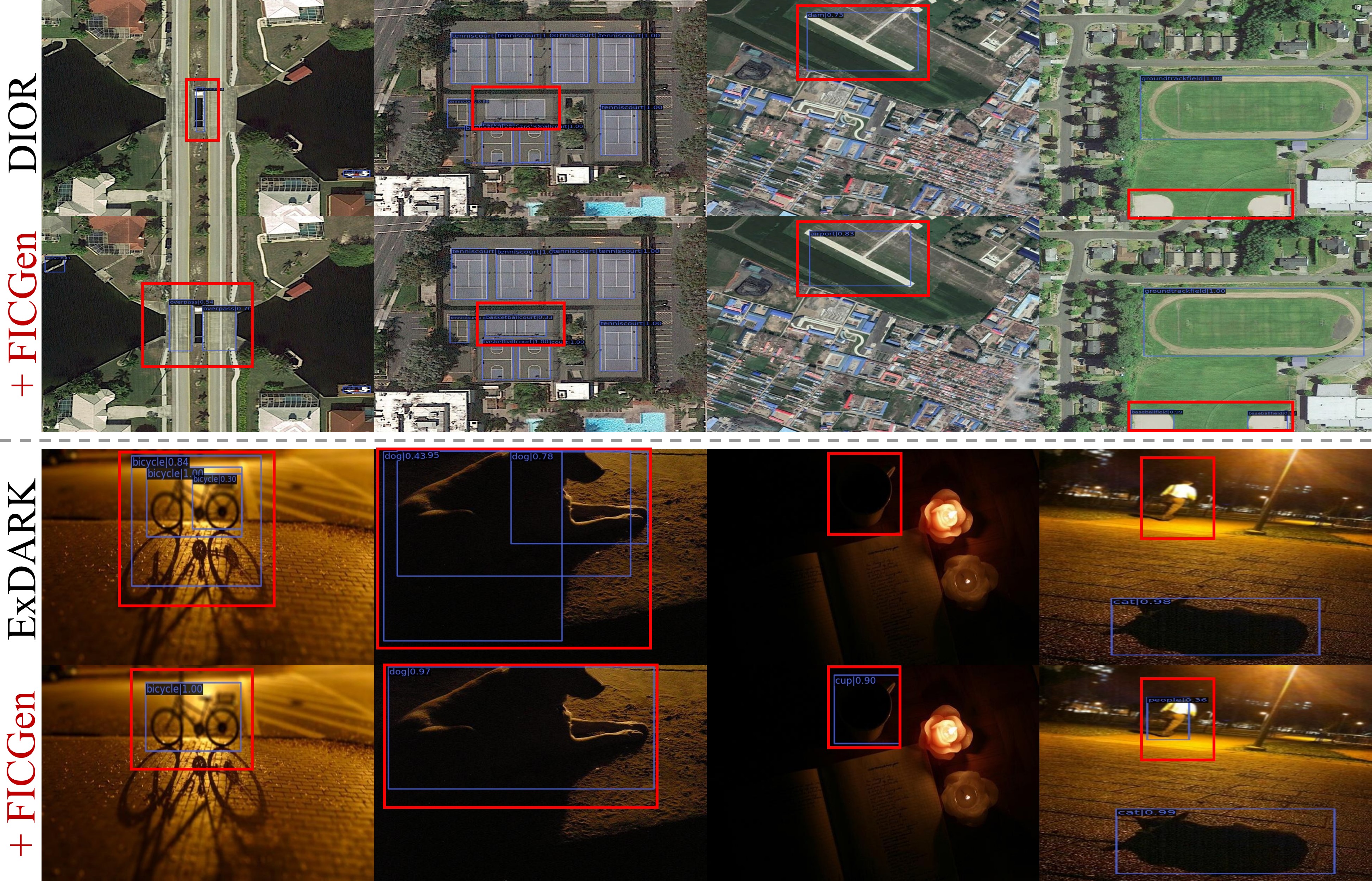}
\caption{Detection results of Faster R-CNN (R50) trained on real images \textit{vs.} real \& FICGen-generated images.}
\label{figdetection}
\end{figure}

\section{Generative Visualizations}
\label{sec4_}
Figs.~\ref{FIG:fulublurvoc}–~\ref{FIG:fuluadverseweather} further showcase the controllable generation capabilities of FICGen, highlighting its ability to address a diverse range of degraded environments, including mild blur, low illumination, underwater, remote sensing, and severe adverse weather conditions. These visualizations emphasize the adaptability and robustness of FICGen in accurately representing the distinctive visual and semantic characteristics of each degraded context. 

In particular, Fig.~\ref{FIG:fulublurvoc} presents synthetic results on the blurred VOC 2012 dataset, illustrating FICGen's capability to handle mild degradations while preserving semantic integrity for categories such as ``\textit{bird}'', ``\textit{cat}'', and ``\textit{sheep}'', without introducing noticeable artifacts or structural inconsistencies. Moreover, even under moderate motion blur, FICGen successfully renders distinguishable shapes for objects like ``\textit{train}'' and ``\textit{horse}'', thereby retaining perceptible fine-grained blurred details for both foreground objects and surrounding backgrounds. 

Figs.~\ref{FIG:fuluexdark} and ~\ref{FIG:fuluunderwater} present that our FICGen excels in replicating the severe conditions of low-illumination and underwater scenarios, while preserving the realism of complex lighting and distortion effects, ultimately delivering semantically aligned and visually realistic degraded samples. Fig.~\ref{FIG:fuluremotesensing} further showcases synthetic results on remote sensing scenes, underscoring FICGen’s capability to generate dense object clusters while preserving coherent spatial relationships between foreground instances and their surroundings. In particular, FICGen accurately renders densely distributed targets such as ``\textit{ship}'' in port areas, while maintaining the correct number of objects. Moreover, it captures contextual consistency by generating ``\textit{vehicles}'' precisely aligned along ``\textit{overpasses}'', ensuring that the synthesized objects seamlessly integrate with the underlying scene structure. 

Fig.~\ref{FIG:fuluadverseweather} presents synthetic samples generated by FICGen under four adverse weather conditions—fog, snow, sand, and rain—on the DAWN dataset, illustrating its strong generalization across visually diverse and challenging degraded scenarios. We can see that FICGen effectively captures the distinctive visual properties of each weather type: \ie, the diffuse scattering and visibility attenuation of ``\textit{fog}''. Beyond instance-level fidelity, FICGen demonstrates the capability to restore critical scene-level elements, such as lane markings and road boundaries, even when they are partially occluded or obscured by rain or sand. Notably, in the ``\textit{sand}'' condition, FICGen generates a highly realistic visual atmosphere, reproducing the chromatic desaturation observed in authentic sandstorm scenarios. 

\section{Limitations and Future Work}
\label{sec5_}
\begin{figure}[!t]
\centering
\includegraphics[width=0.46\textwidth]{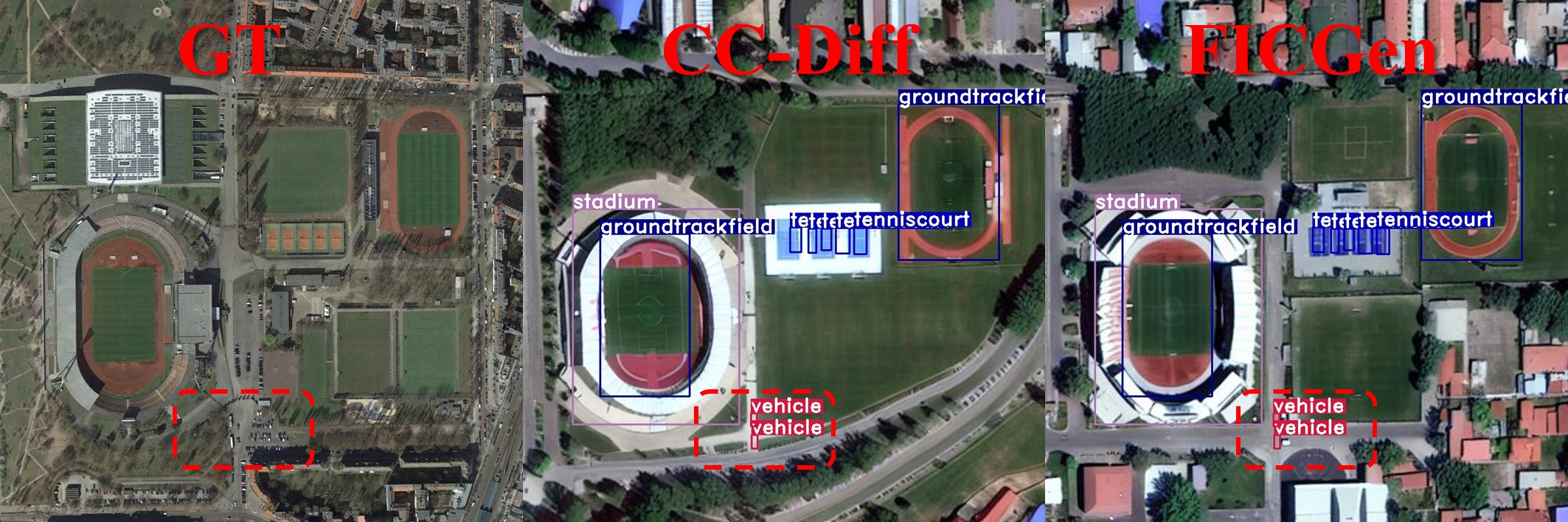}
\caption{Failure cases of FICGen, where red dashed boxes denote the missing instances.}
\label{figbadcases}
\end{figure}

\begin{figure*}
\centering
\includegraphics[width=1.00\textwidth]{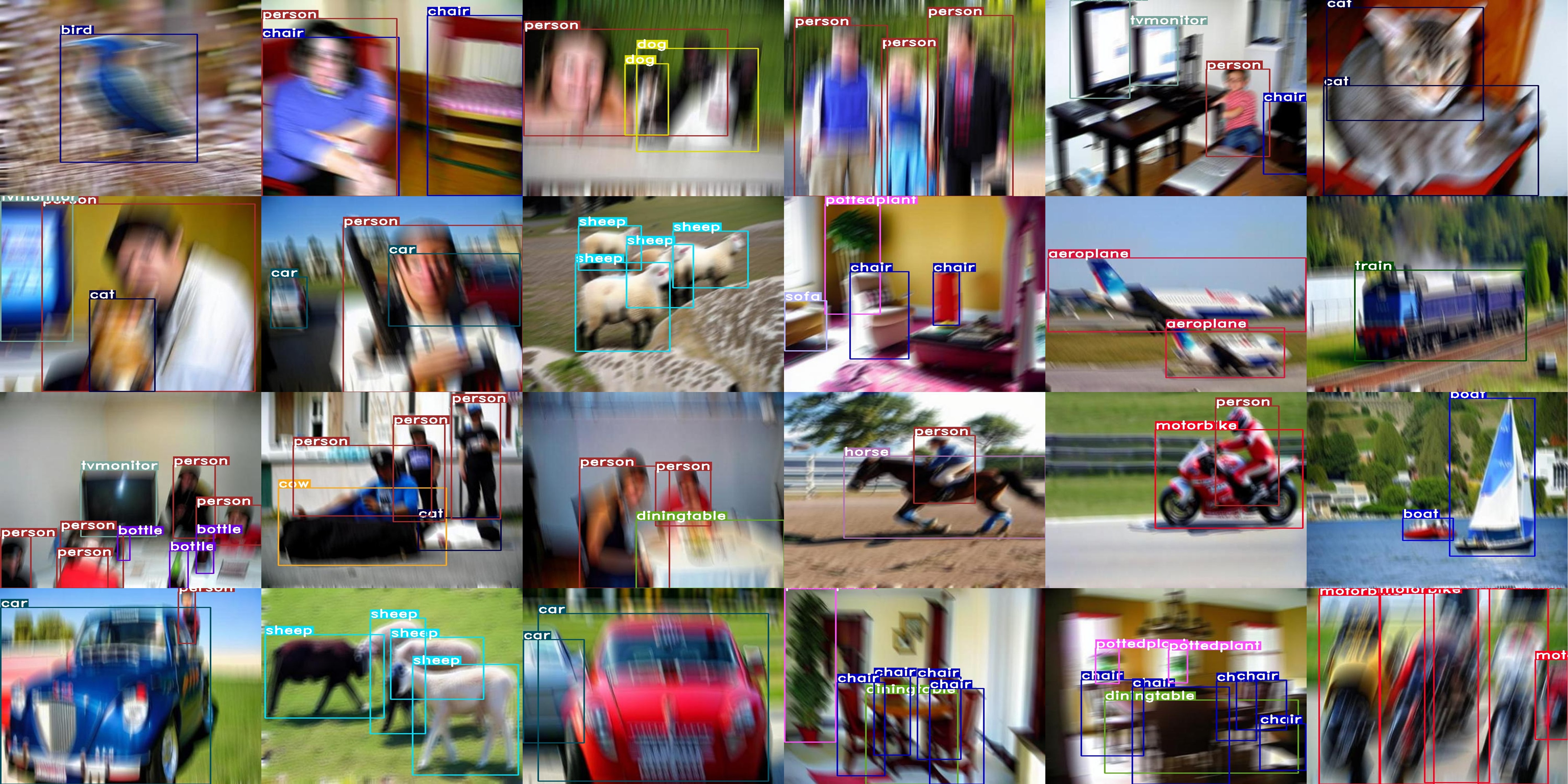}
\caption{Visualization results generated by our FICGen under the mild blur (blurred VOC 2012), with geometric layouts and corresponding object categories superimposed on the generated images.}
\label{FIG:fulublurvoc}
\end{figure*}

\begin{figure*}
\centering
\includegraphics[width=1.00\textwidth]{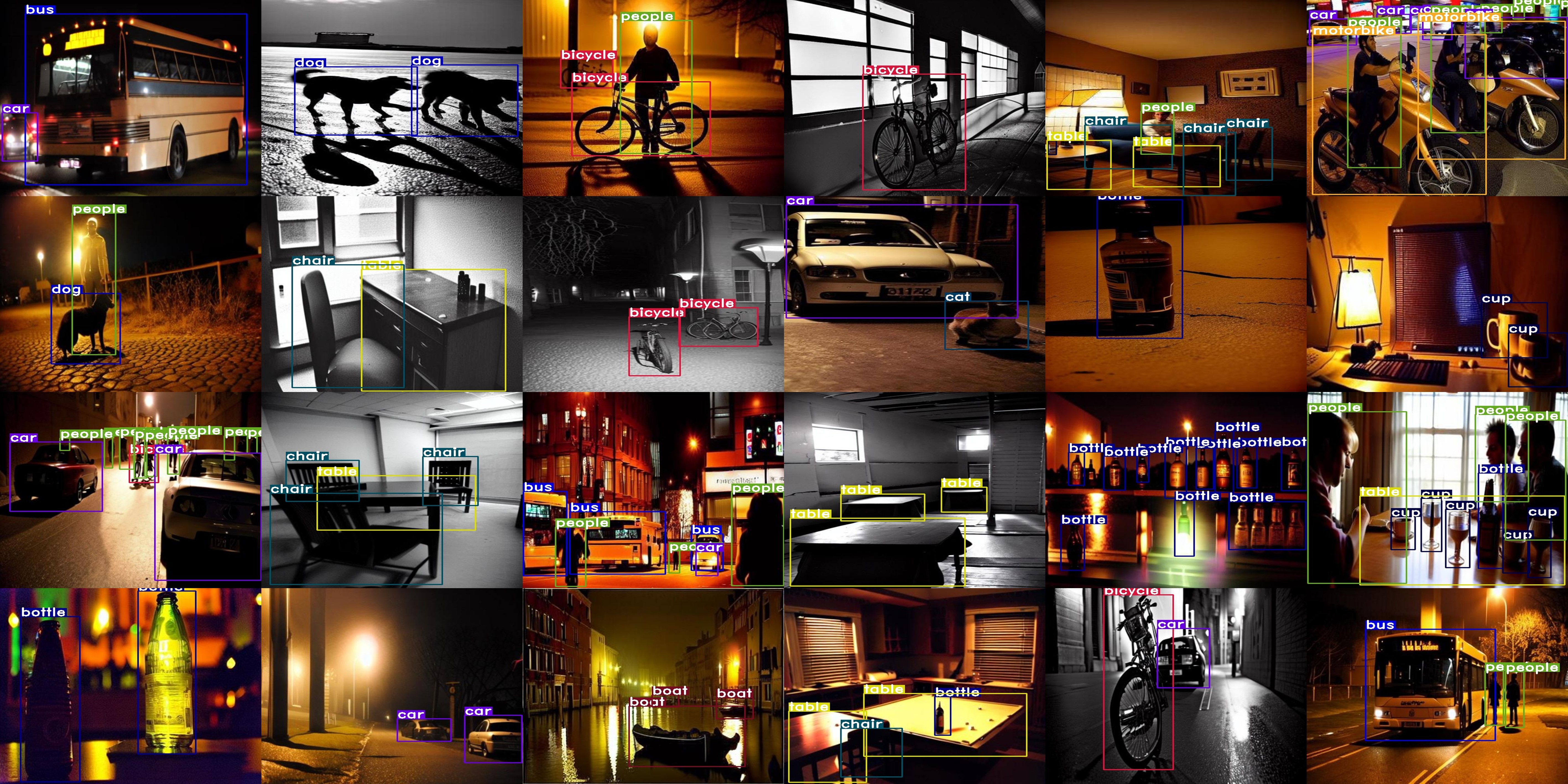}
\caption{Visualization results generated by our FICGen under low-light conditions (ExDARK~\cite{loh2019getting}), with geometric layouts and corresponding object categories superimposed on the generated images.}
\label{FIG:fuluexdark}
\end{figure*}

\begin{figure*}
\centering
\includegraphics[width=1.00\textwidth]{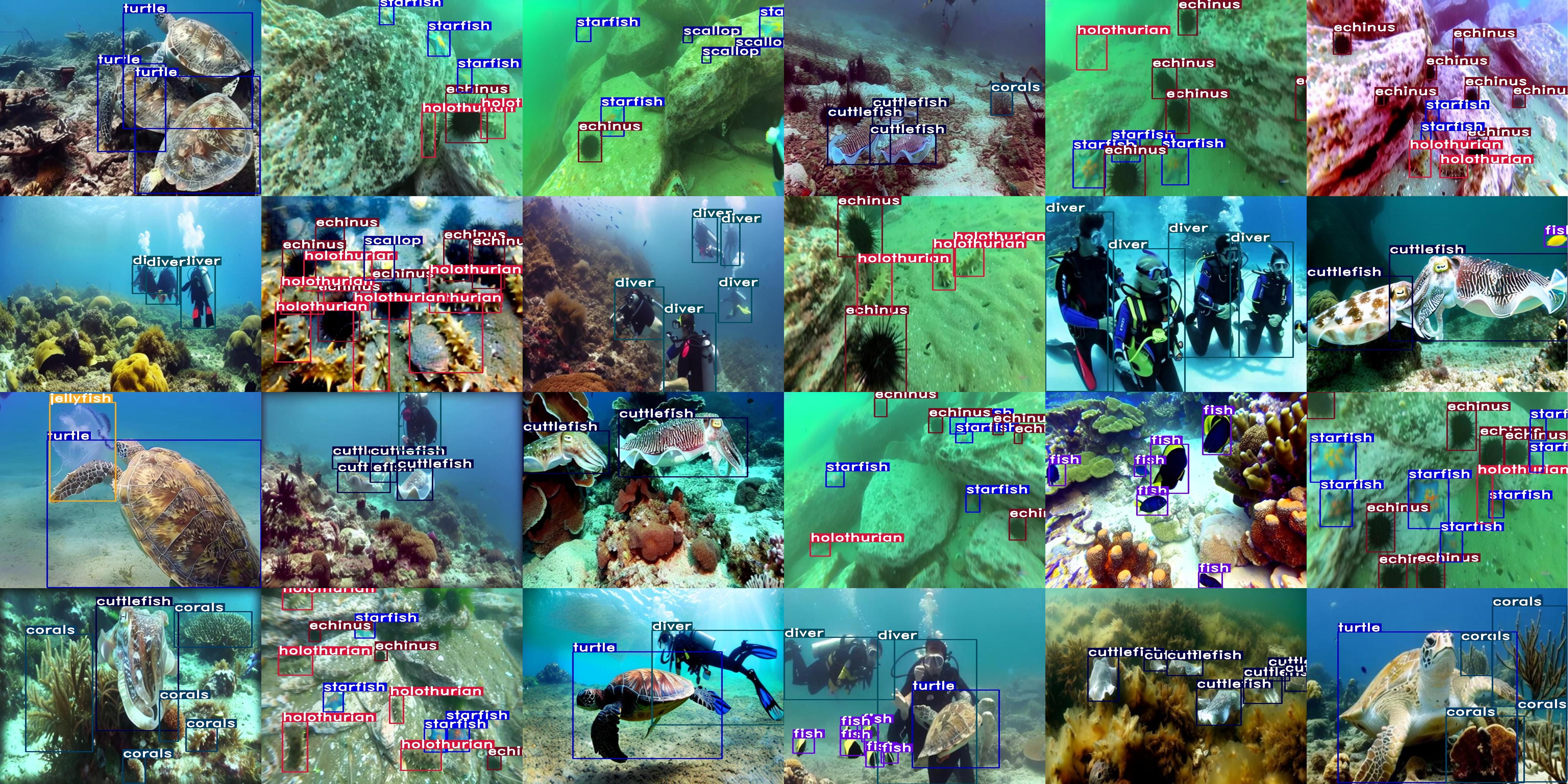}
\caption{Visualization results generated by our FICGen under the underwater scene (RUOD~\cite{fu2023rethinking}), with geometric layouts and corresponding object categories superimposed on the generated images.}
\label{FIG:fuluunderwater}
\end{figure*}

\begin{figure*}
\centering
\includegraphics[width=1.00\textwidth]{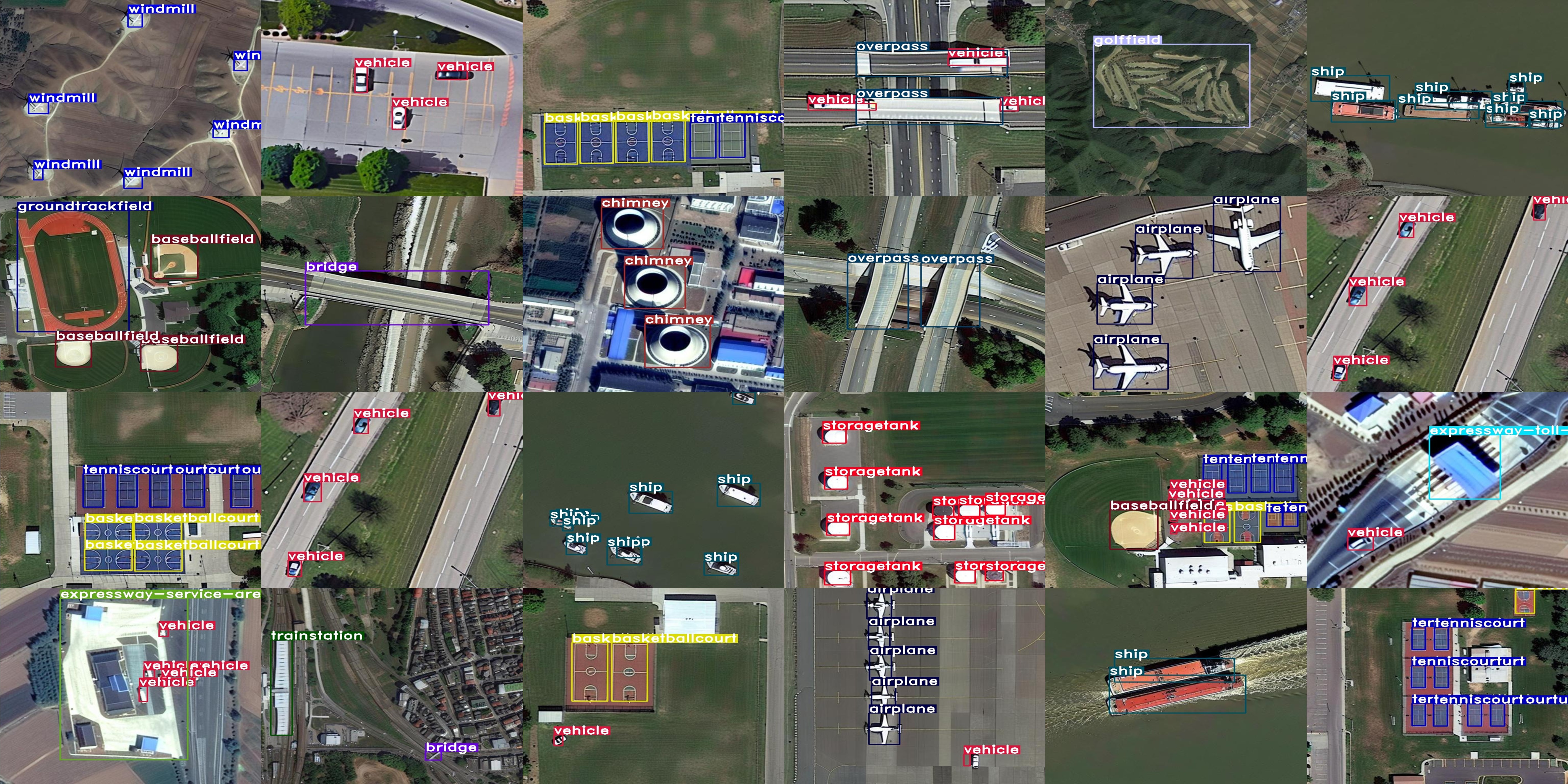}
\caption{Visualization results generated by our FICGen under the remote sensing scene (DIOR~\cite{li2020object}), with geometric layouts and corresponding object categories superimposed on the generated images.}
\label{FIG:fuluremotesensing}
\end{figure*}

\begin{figure*}
\centering
\includegraphics[width=1.00\textwidth]{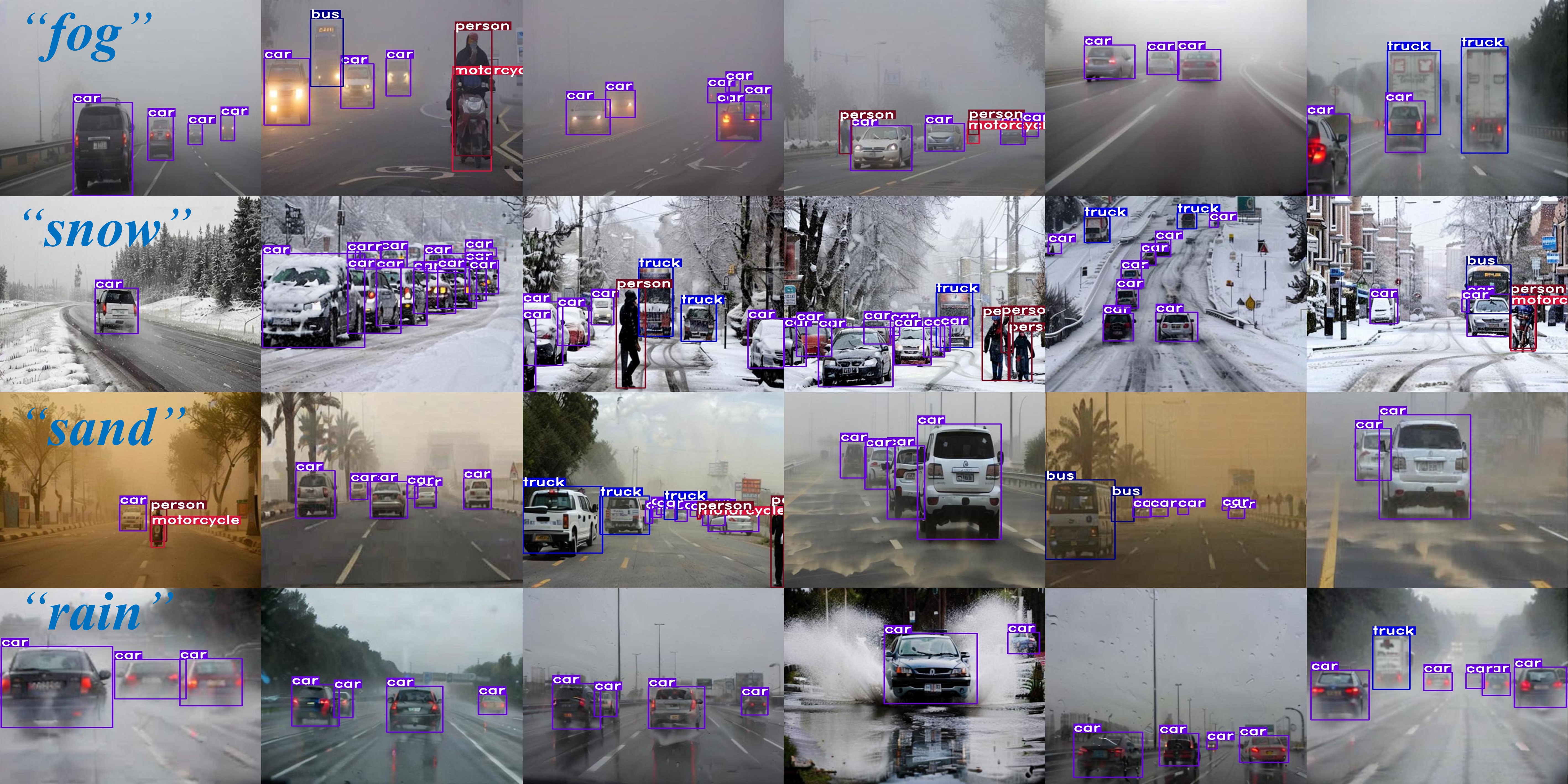}
\caption{Visualization results generated by our FICGen under the adverse weather condition (DAWN~\cite{kenk2020dawn}), with geometric layouts and corresponding object categories superimposed on the generated images.}
\label{FIG:fuluadverseweather}
\end{figure*}

Fig.~\ref{figbadcases} illustrates the failure case, where FICGen struggles to synthesize high-resolution remote sensing images with precise representations of small objects such as ``\textit{vehicles}.'' This limitation primarily arises from the inherent downsampling operations in latent diffusion models, which suppress fine-grained structural details. Future work will focus on extending the frequency-inspired paradigm to 3D content generation, such as camouflaged video synthesis, and exploring richer control modalities beyond bounding boxes, including semantic masks, to enhance the precision and controllability of contextual generation in degraded scenarios.

\end{document}